\documentclass[letterpaper, 10 pt, conference]{ieeeconf}  

\IEEEoverridecommandlockouts                              

\usepackage{graphicx} 
\usepackage{amsmath}
\usepackage{amssymb}
\usepackage{mathtools}
\usepackage{lettrine}
\usepackage{type1cm}
\usepackage{threeparttable}
\usepackage{caption}
\usepackage{subcaption}
\usepackage{ragged2e}
\usepackage{color}
\usepackage{siunitx}
\usepackage{bm}
\usepackage{hyperref}
\usepackage{pifont}
\usepackage{cite}

\hypersetup{
    colorlinks=true,
    linkcolor=blue,
    citecolor=blue, 
    urlcolor=blue
}

\title{\LARGE \bf
Adaptive Perching and Grasping by Aerial Robot\\
with Light-weight and High Grip-force Tendon-driven\\
Three-fingered Hand using Single Actuator
}

\author{Hisaaki Iida$^{1}$, Junichiro Sugihara$^{2}$, Kazuki Sugihara$^{2}$, Haruki Kozuka$^{2}$, \\ Jinjie Li$^{1}$, Keisuke Nagato$^{1}$, and Moju Zhao$^{1}$%
  \thanks{Hisaaki Iida (corresponding author), Jinjie Li, Keisuke Nagato, and Moju Zhao are with the Department of Mechanical-Engineering, The University of Tokyo, Bunkyo-ku, Tokyo 113-8656, Japan (e-mail: {\tt\small \{iida, jinjie-li, chou\}@dragon.t.u-tokyo.ac.jp nagato@hnl.t.u-tokyo.ac.jp).}}
  \thanks{Junichiro Sugihara, Kazuki Sugihara, and Haruki Kozuka are with the Department of Mechano-Infomatics, The University of Tokyo, Bunkyo-ku, Tokyo 113-8656, Japan (email: {\tt\small \{j-sugihara, sugihara, kozuka\}@jsk.imi.i.u-tokyo.ac.jp).}}
}

\LettrineRealHeighttrue

\begin{document}

\maketitle
\thispagestyle{empty}
\pagestyle{empty}


\begin{abstract}

Aerial robots, especially multirotor type, have been utilized in various scenarios such as inspection, surveillance, and logistics. The most critical issue for multirotor type is the limited flight time due to the large power consumption to hover against gravity. 
Inspired by nature, various research areas focus on the perching and grasping ability by deploying a gripper on the multirotor to grasp arboreal environments to save energy; however, most of the mechanical design for gripper restricts the approach path, significantly limiting the performance of perching and grasping. In addition, it is also challenging to design a light gripper that also offers sufficiently large grip force to hang itself. Therefore, in this work, we develop a single-actuator hand for aerial robot that enables adaptive grasping of various objects, and thus can perch from various approach directions. First, we present the design of the lightweight three-fingered hand with a pair of special two-dimensional differential plates that enables adaptive grasping with a single actuator. In addition, we develop a unique control method for the over-actuated aerial robot equipped with this hand to perform both adaptive pendulum-like perching and detachment. Finally, we demonstrate the feasibility of the prototype hand via load bearing and object grasping experiments, along with in-flight perching experiments. 

\end{abstract}


\section{I\footnotesize NTRODUCTION}
\label{sec:introduction}

Today, aerial robots have gained significant maneuverability and are being used to perform tasks that are impossible for humans or non-flying machines, for example, monitoring large structures from various perspectives, environmental exploration, or disaster rescue\cite{observation, mapping, disaster}. Along with those extended operations, various types of perching mechanisms such as the below-body type\cite{underhandperching, underhandperching2, underhandperching3, underhandperching4} or the overhead type\cite{overheadperching, overheadperching2} have been developed as countermeasures against the fundamental weakness of aerial robots: their high power consumption. However, in current research, most perching units are placed directly above or below the robot body, resulting in perching approach paths that are limited to directions nearly vertical to the target object. This limitation reduces the selection of attachment positions, especially near walls, where the robot body is prone to interference or obstacles are present in the surroundings. Several studies are proposing horizontal perching using fixed-wing aerial robots \cite{horizon, horizon2}, but generally suffer from low maneuverability in the air, making them particularly impractical for use in an intricate environment.
\begin{figure}[tbp]
    \centering
    \begin{subfigure}[tb]{0.37\linewidth}
        \centering
        \includegraphics[keepaspectratio, width=\linewidth]{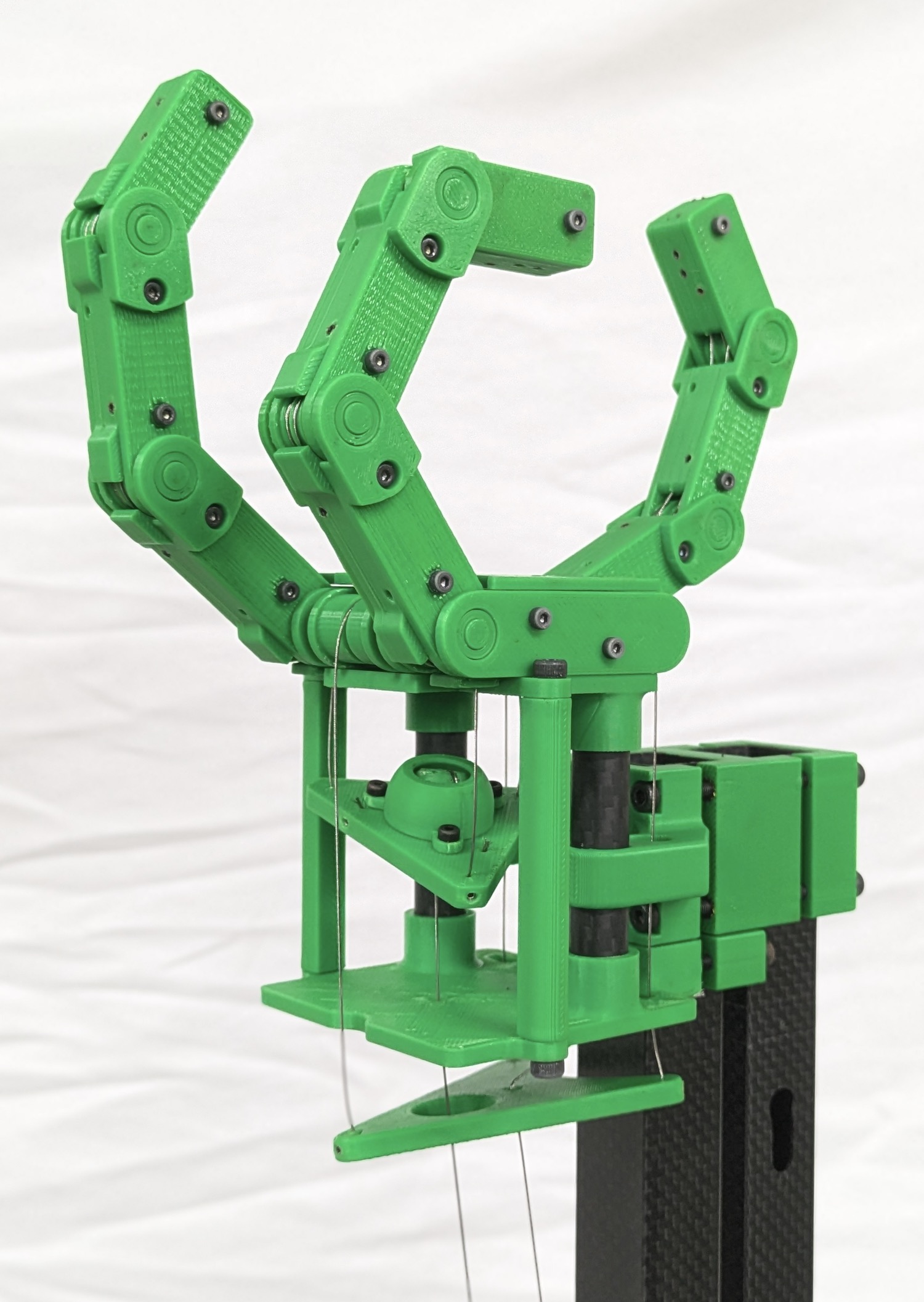}
        \caption{}
        \label{firstfiga}
    \end{subfigure}
    \hfill
    \begin{subfigure}[tb]{0.61\linewidth}
    \centering
        \includegraphics[keepaspectratio, width=\linewidth]{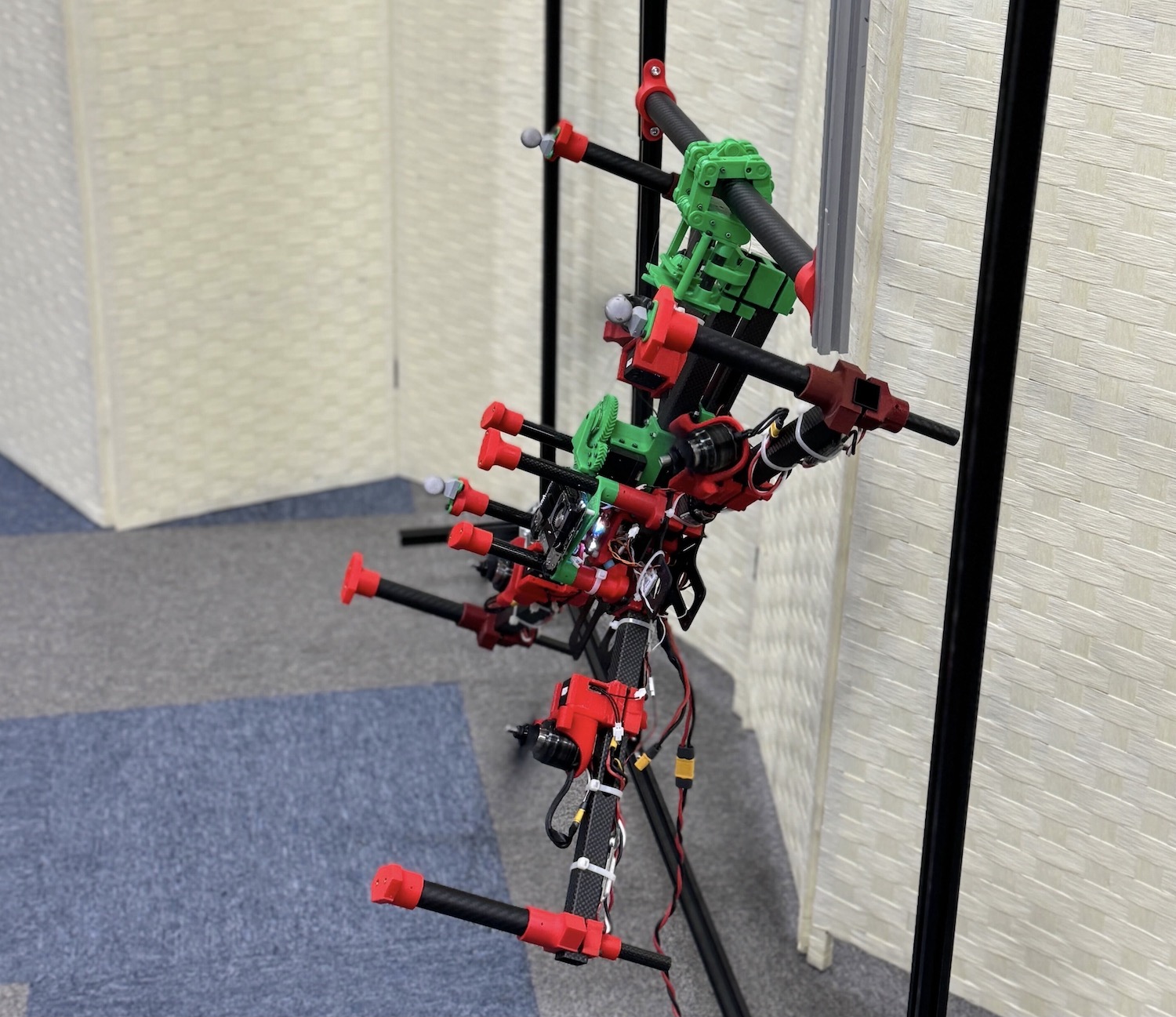}
        \caption{}
        \label{firstfigb}
    \end{subfigure}
    \captionsetup{justification=raggedright, singlelinecheck=false}
    \caption{\quad\textbf{Tri-force hand}: The robot hand module for perching from multi-direction and adaptive grasping. \textbf{(a)} Hand module for adaptive grasping and multi-directional perching. \textbf{(b)} Appearance of the quadrotor equipped with the proposed hand performing vertical takeoff.}
    \label{firstfig}
    \vspace{-3mm}
\end{figure}
Furthermore, most existing perching units lack any functionality other than attachment, resulting in unnecessary weight during flight. Providing additional functionality for these units leads to improved power efficiency. Therefore, this research proposes a new lightweight end effector for aerial robot capable of multi-directional perching and grasping objects adaptively and firmly, driven by a single actuator, as shown in Fig. \ref{firstfig}.

With regard to design, the most common method of perching involves a simple robotic hand. This method can be further classified according to the driving mechanisms, including springs\cite{ultra}, tendon-driven mechanisms\cite{armed, tendondriven}, and gravity-driven mechanisms\cite{SNAG, selfmass, selfmass2}. However, gravity-driven mechanisms strongly limit the direction of attachment and make them unsuitable for multi-directional perching, while using springs makes it difficult to control the force and the angle of the finger actively. In contrast, tendon-driven mechanisms overcome the aforementioned drawbacks and offer several advantages, such as relatively high mechanical compliance upon contact with objects and greater freedom in positioning the actuator to minimize the rotational moment of the aerial robot. Therefore, in this research, we select a tendon-driven robotic hand as the end effector. 
\begin{figure}[!b]
    \centering
    \begin{subfigure}[tb]{0.24\linewidth}
        \centering
        \includegraphics[keepaspectratio, width=\linewidth]{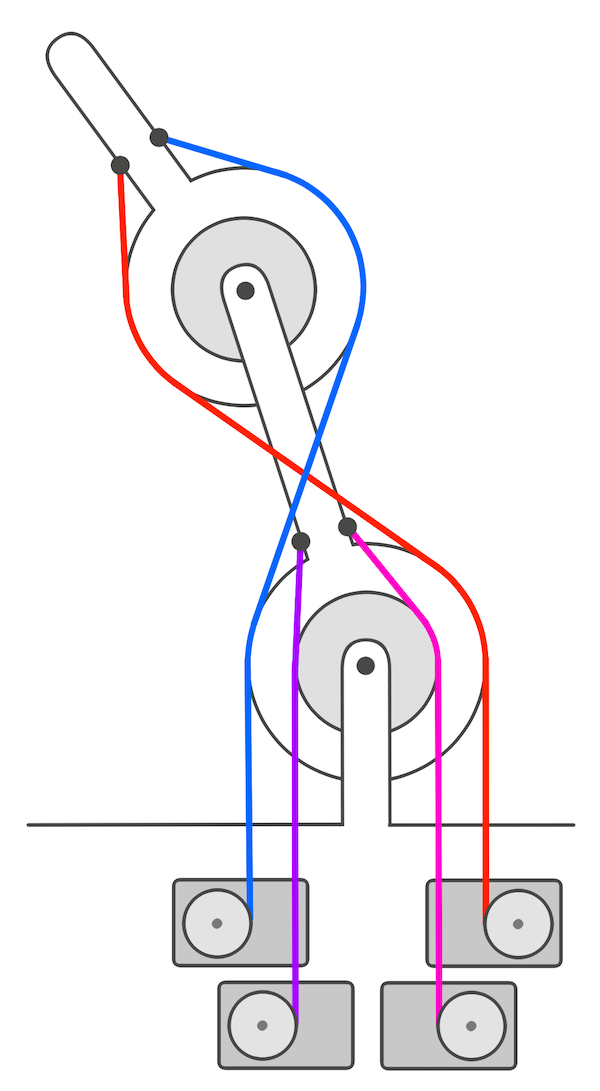}
        \caption{CF-TDM}
        \label{cftdm}
        \hspace{1mm}
    \end{subfigure}
    \begin{subfigure}[tb]{0.24\linewidth}
    \centering
        \includegraphics[keepaspectratio, width=\linewidth]{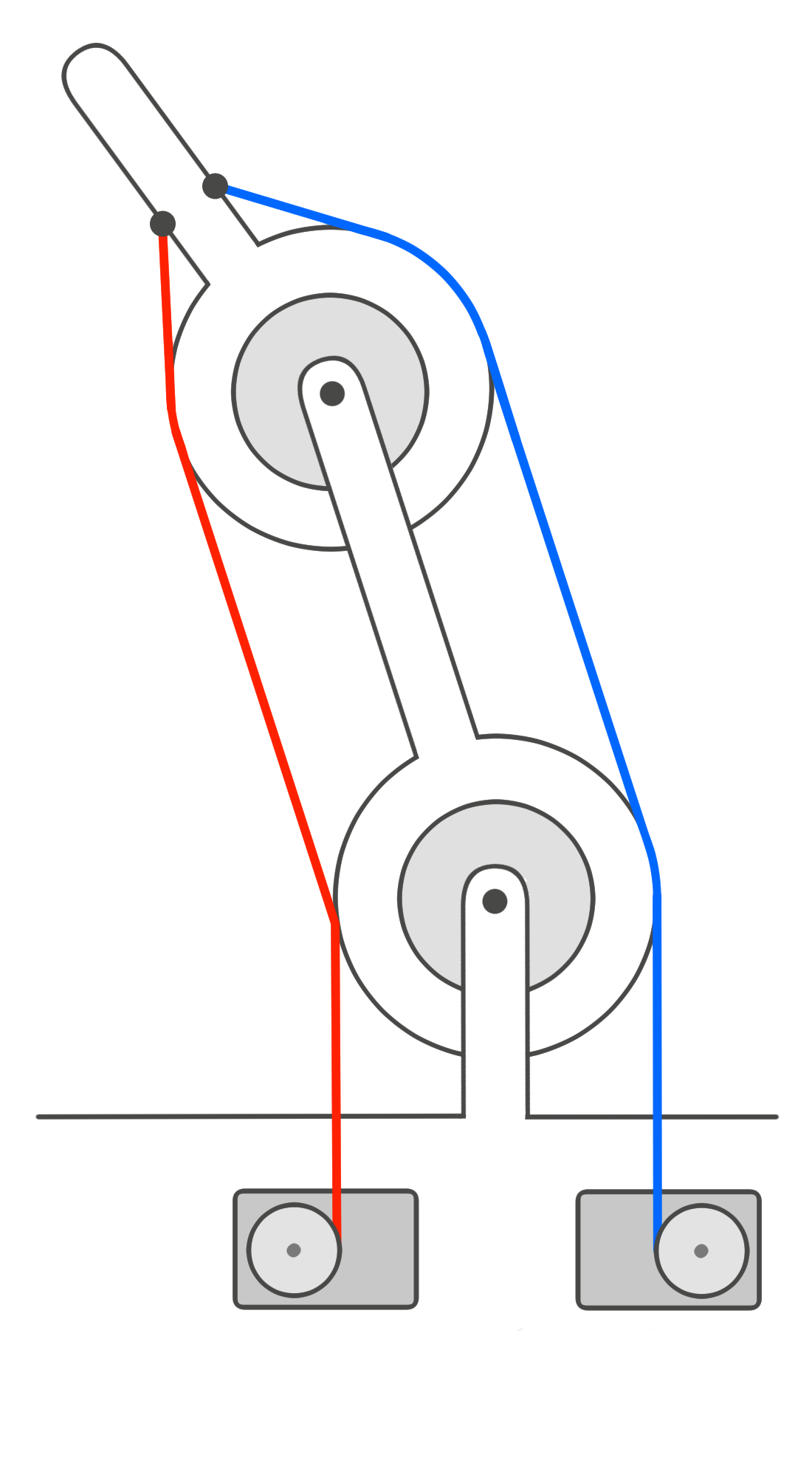}
        \caption{UF-TDM}
        \label{uftdm}
        \hspace{1mm}
    \end{subfigure}
    \begin{subfigure}[tb]{0.24\linewidth}
    \centering
        \includegraphics[keepaspectratio, width=\linewidth]{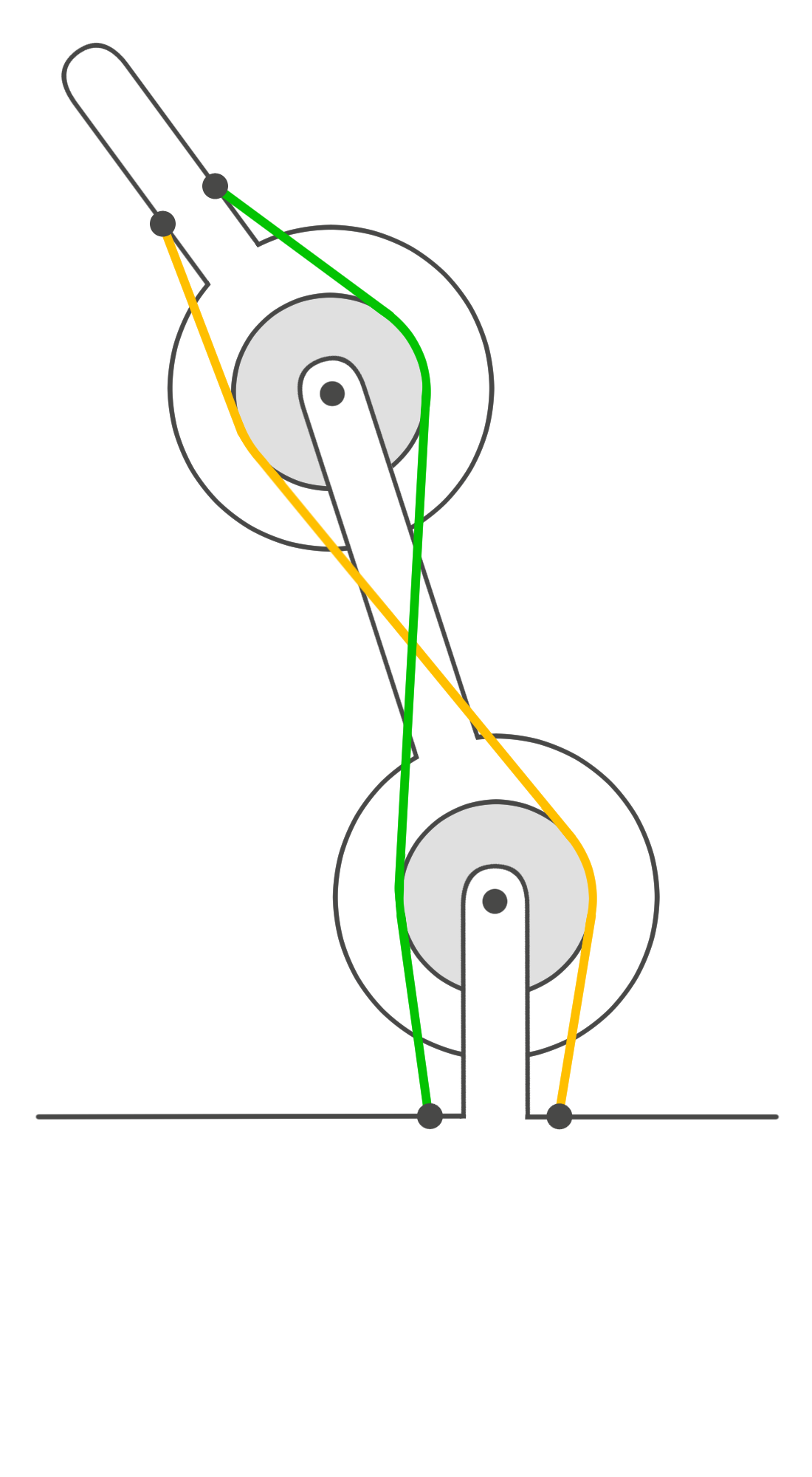}
        \caption{UP-TDM}
        \label{uptdm}
        \hspace{1mm}
    \end{subfigure}
    \begin{subfigure}[tb]{0.24\linewidth}
    \centering
        \includegraphics[keepaspectratio, width=\linewidth]{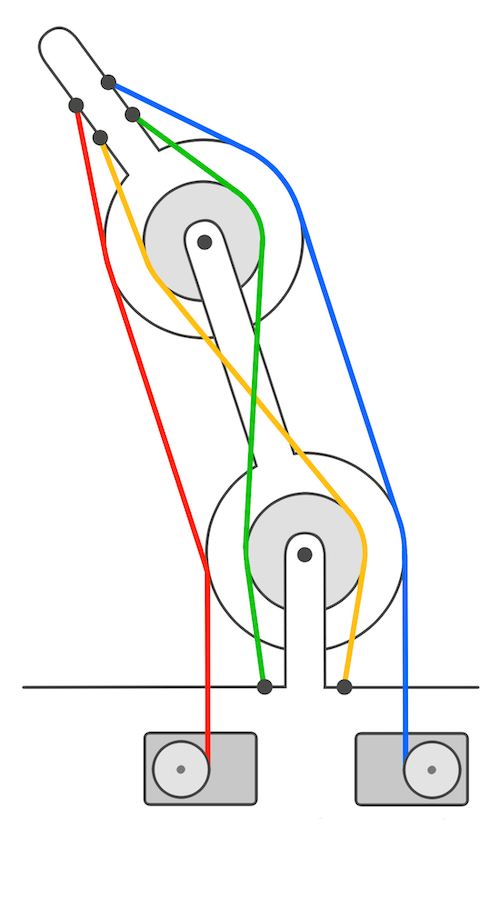}
        \caption{CS-TDM}
        \label{cstdm}
        \hspace{1mm}
    \end{subfigure}
    \begin{subfigure}[tb]{0.99\linewidth}
        \centering
        \includegraphics[keepaspectratio, width=\linewidth]{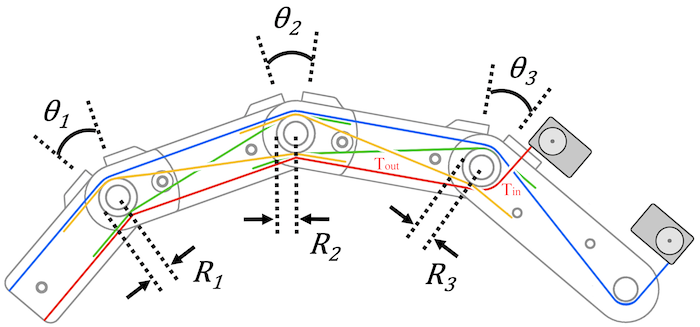}
        \caption{}
        \label{linkmodel}
        \hspace{1mm}
    \end{subfigure}
    \captionsetup{justification=raggedright, singlelinecheck=false}
    \caption{\quad \textbf{Finger tendon arrangement.} \textbf{(a)} Diagram of tendon arrangement in tendon-driven systems; CF-TDM. \textbf{(b)(c) (d)} general UF-TDM, UP-TDM and CS-TDM model. \textbf{(e)} Three-joint linkage finger model and Tendon arrangement for synchronizing the angles of adjacent joints. The classification and naming are based on \cite{Ozawa}.}
    \vspace{-3mm}
\end{figure}

In general, the actuator accounts for a large percentage of the overall system weight, thus it is essential to minimize the number of actuators while maintaining controllability. Since controllable full tendon-driven mechanism (CF-TDM\cite{Ozawa}) requires at least as many actuators as the number of joints, we opt for a controllable semi tendon-driven mechanism (CS-TDM\cite{Ozawa}) that requires two actuators for each finger. However, further reduction in the number of actuators is required for integration into aerial robots.

Therefore, in this research, we extend the existing study of a straight differential mechanism\cite{differential} to the opposing three-fingered hand, using triangular differential plates that allow adaptive grasping to be achieved while reducing the total number of actuators to just one.

Regarding control, since the perching approach proposed in this research is performed from various directions and altitude, the base aerial robot must be able to stably change its tilt angle and approach while maintaining attitude. To meet this requirement, we used an overactuated quadrotor\cite{j-sugihara} as the base robot in this study. During the perching sequence, the robot body rotates around an axis as shown in Fig. \ref{firstfigb}, which we refer to as ``Pendulum perching". For this special maneuver, we have devised a specialized trajectory planning method about thrust control during landing and position control during takeoff.\\

The main contributions of this work can be summarized as follows.

\begin{enumerate}

    \item[1)] We propose the mechanical design of the lightweight hand for the aerial robot capable of perching and adaptive grasping, using a tendon-driven mechanism and a singular actuator with two-dimensional differentials.
    \item[2)] We describe a motion planning method for pendulum perching from multiple directions, especially from the side, and takeoff for detachment from a vertical state.
    \item[3)] We conducted experiments of pendulum perching using actual aerial robot equipped with the hand.
    
\end{enumerate}

The remainder of this paper is structured as follows.

Section \ref{sec:design} introduces the mechanical design of the hand. Section \ref{sec:control} explains the robot control, while Section \ref{sec:expandres} presents the experimental procedures and results for validation. Finally, Section \ref{sec:conclusion} concludes the paper.


\section{D\footnotesize ESIGN}
\label{sec:design}

In this section, we introduce the base aerial robot design, the mechanical design of the hand capable of perching and adaptive grasping. The key aspects of this structure are the finger model using CS-TDM and triangular plates as two-dimensional differentials.

\subsection{Base Aerial Robot Model}

To achieve stable perching from multiple directions, this study employs an 8 degree-of-freedom over actuated quadrotor as the base robot for mounting the hand. This quadrotor utilizes tilt rotors to ensure the degrees of freedom while allowing for linear control. The details of the control are explained in Section \ref{sec:control}.

\subsection{Three-Fingered Model}

CF-TDM is one of the general models of tendon-driven hands. It allows for the construction of a precise musculoskeletal model using multiple actuators; Fig. \ref{cftdm} shows \textit{2N} model, requires two actuators for one joint. For weight reduction, we use another model CS-TDM shown in Fig. \ref{cstdm}, constructed using a combination of UF-TDM (Fig. \ref{uftdm}) and UP-TDM (Fig. \ref{uptdm}). As shown in Fig. \ref{linkmodel}, the hand in this study uses a human skeletal structure. Three links corresponding to the metacarpal bones are linked by 4 passive tendons arranged in an X shape represented by yellow and green lines and driven by the active flexor / extensor tendon represented by a red / blue line. Since both passive and active tendons are used, the kinematics of the CS-TDM can be described as follows\cite{Ozawa}:
\begin{equation}
    \begin{bmatrix}
        \dot{\bm{l_a}}\\
        \dot{\bm{l_p}}
    \end{bmatrix}
    =
    \bm{J}\dot{\bm{\theta}} + \bm{B}\dot{\bm{\phi}}
    =
    \begin{bmatrix}
        \bm{J_a}\\
        \bm{J_p}
    \end{bmatrix}
    \dot{\bm{\theta}} + 
    \begin{bmatrix}
        \bm{B_a}\\
        \bm{0}
    \end{bmatrix}
    \dot{\bm{\phi}},
    \label{eq:ozawa}
\end{equation}
where $\bm{l}$ is tendon elongation matrix, $\bm{\theta}$ is joint angle matrix, $\bm{\phi}$ is servo angle matrix, $\bm{J}$ is jacobian between $\bm{\dot{l}}$ and $\bm{\dot{\theta}}$, and $\bm{B}$ is jacobian between actuators and tendons. Both $\bm{J}$ and $\bm{B}$ are constant matrices that depend on the joint pulley diameter and actuator pulley diameter, respectively. 

Therefore by organizing the lower rows of equation (\ref{eq:ozawa}) which represents passive tendons, we get $\bm{J_p \dot{\theta}} = 0$, can be rephrased as $\bm{J_p \theta} = 0$ where $\dot{\bm{l}} = 0$. In proposed hand, $\bm{J_p \theta}$ can be simplified with following equations:
\begin{equation}
    \bm{J_p} \bm{\theta} = 
    \begin{bmatrix}
        R_1 & -R_2 & 0 \\
        -R_1 & R_2 & 0 \\
        0 & R_2 & -R_3 \\
        0 & -R_2 & R_3 \\
    \end{bmatrix}
    \bm{\theta} = 
    \begin{bmatrix}
        R_1 & 0 \\
        -R_1 & 0 \\
        0 & R_1 \\
        0 & -R_1 \\
    \end{bmatrix}
    \begin{bmatrix}
        \theta_1 - \theta_2 \\
        \theta_2 - \theta_3 \\
    \end{bmatrix},
    \label{rtheta}
\end{equation}
where $R_i$ represents the radius of the $i$-th joint pulley, assumed that all are the same here. Given that $R_1$ is not 0, (\ref{rtheta}) indicates that $\theta_1, \theta_2$ and $\theta_3$ must have same value, thus the angles of the all joints are kept same.

\begin{figure}[tbp]
    \centering
    \begin{subfigure}[b]{\linewidth}
        \centering
        \includegraphics[keepaspectratio, width=0.95\linewidth]{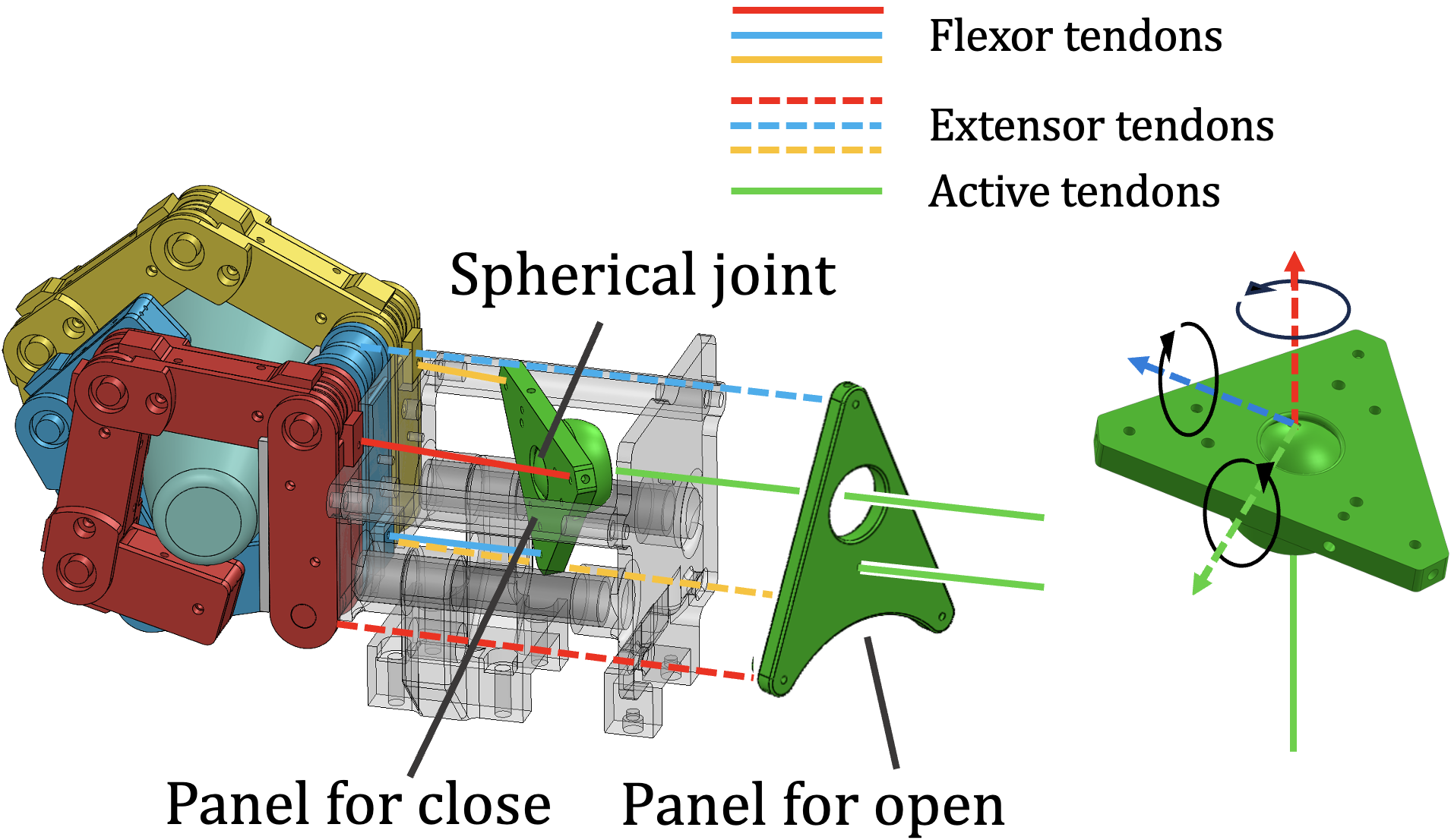}
        \caption{}
        \label{triforcelook}
        \vspace{-2mm}
        \hspace{1mm}
    \end{subfigure}
    \hfill
    \begin{subfigure}[t]{0.49\linewidth}
    \centering
        \includegraphics[keepaspectratio, width=\linewidth]{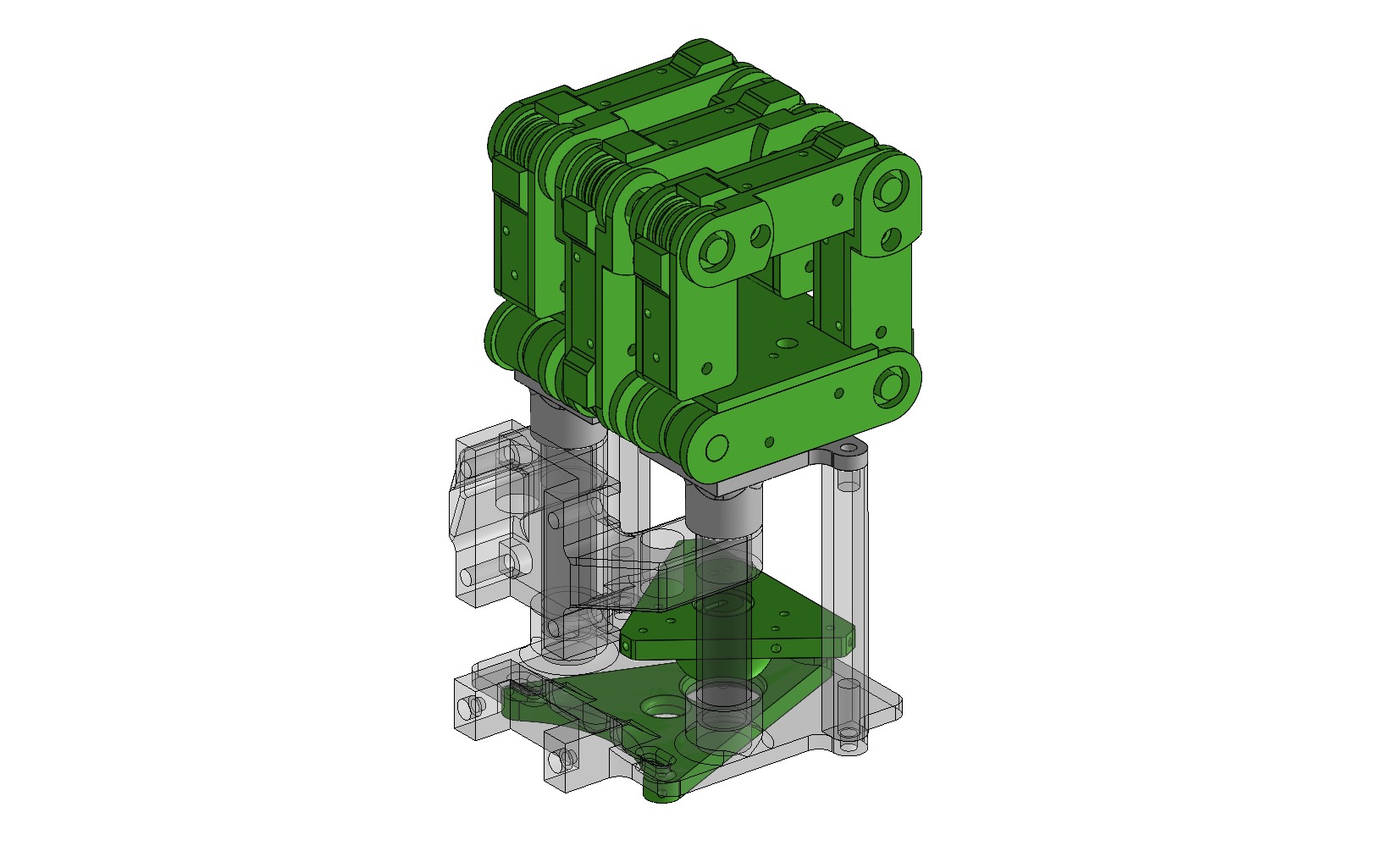}
        \caption{}
        \label{triforceclose}
        \vspace{-5mm}
        \hspace{1mm}
    \end{subfigure}
    \begin{subfigure}[t]{0.49\linewidth}
    \centering
        \includegraphics[keepaspectratio, width=\linewidth]{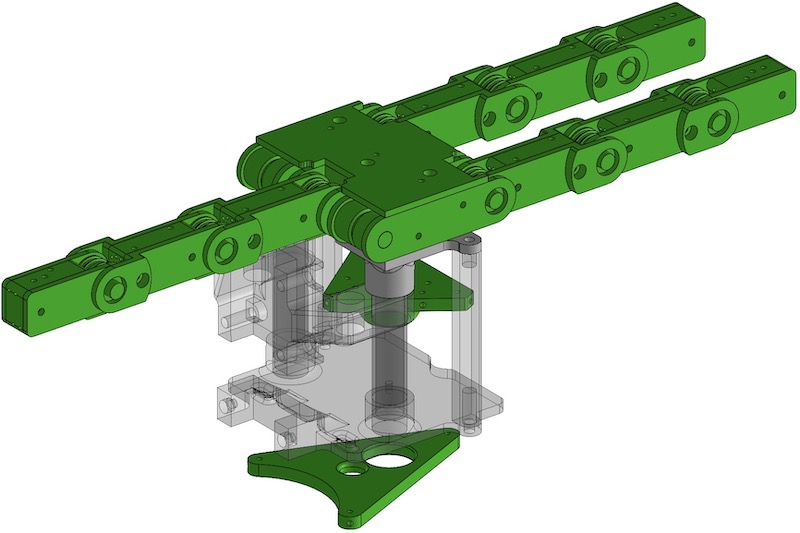}
        \caption{}
        \label{triforceopen}
    \end{subfigure}
    \captionsetup{justification=raggedright, singlelinecheck=false}
    \caption{\quad \textbf{Mechanical design of the ``Tri-force system".} \textbf{(a)} The posture of differential plates when grasping a truncated cone-like shape. \textbf{(b)(c)} Fully closed/open state of the hand.}
    \vspace{-3mm}
\end{figure}

 With this tendon arrangement, the joint angles of each finger are geometrically constrained and all the force exerted by the actuator is used for gripping the object. Consequently, it is possible to transition between a flat state and a gripping state, as shown in Fig. \ref{triforceclose} and \ref{triforceopen}, while exerting a strong gripping force with only one actuator per finger. Additionally, by wrapping wires around built-in joint pulley, the flexor tendons can exert even greater tension due to friction as described by \textit{Euler's belt formula}\cite{eulerbelt}:
\begin{equation}
    T_{out} = T_{in}\,e^{\mu\delta},
    \label{eq:euler}
\end{equation}
where $\mu$ is coefficient of friction and $\delta$ is wire wrapping angle, tension $T_{in}$ and $T_{out}$ are shown in FIg. \ref{linkmodel}. For the value of the static friction coefficient $\mu$, we assumed 0.3\cite{friction}. Moreover, considering the negative effect of additional weight during flight, our objective is to further reduce the number of actuators in our hand design.

\subsection{Two-Dimensional Differential Plate}
Negin Nikafrooz et al. proposed a mechanism that reduces the number of actuators in the parallel five-fingered hand by using a straight 1-dimensional differential \cite{differential}. However, in their research, the movement of one finger surely affects the others, resulting in low adaptability, especially to objects with shape that varies significantly in diameter or concave features. 

In this research, we extend this mechanism to an opposing three-fingered design, proposing a structure that ensures higher adaptability. As shown in Fig. \ref{triforcelook}, three flexor tendons and three extensor tendons extended to the back of the hand are connected to the vertices of triangular plates. These plates function as two-dimensional differentials by tilting according to the movement of the tendons, while preventing each finger's movement from affecting the others.

As the specific vertex of the panel for closing hand is pulled to the right of Fig. \ref{triforcelook}, the corresponding finger bends, and the corresponding vertex of the panel for opening hand is pulled to the left. The reverse occurs in the opposite situation. By connecting these plates with a single active tendon driven by single actuator, the hand can grasp complex-shaped objects with only one actuator that the hands using previous differentials could not handle. Adaptability of the hand is conducted in Section \ref{sec:expandres}.

To avoid applying bending stress to the wire at the connection point between the tendon and the plate, a spherical joint is introduced in the central part of the plate for closure. Since the differential plate is almost an equilateral triangle, this structure ensures that the forces acting on each finger are symmetrical and nearly equal, allowing the hand to adapt to and grasp objects. 

\subsection{Maximum Grip Force for Perching}
\begin{figure}[tbp]
    \centering
    \begin{subfigure}[b]{0.32\linewidth}
        \centering
        \includegraphics[keepaspectratio, width=\linewidth]{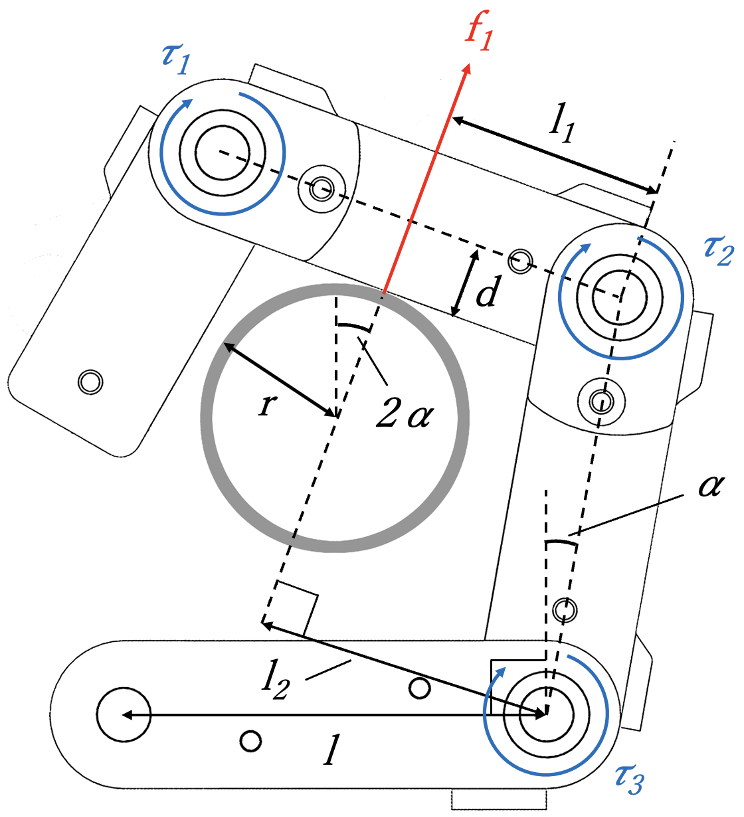}
        \caption{}
        \label{lessalpha}
    \end{subfigure}
    \hfill
    \begin{subfigure}[b]{0.62\linewidth}
    \centering
        \includegraphics[keepaspectratio, width=\linewidth]{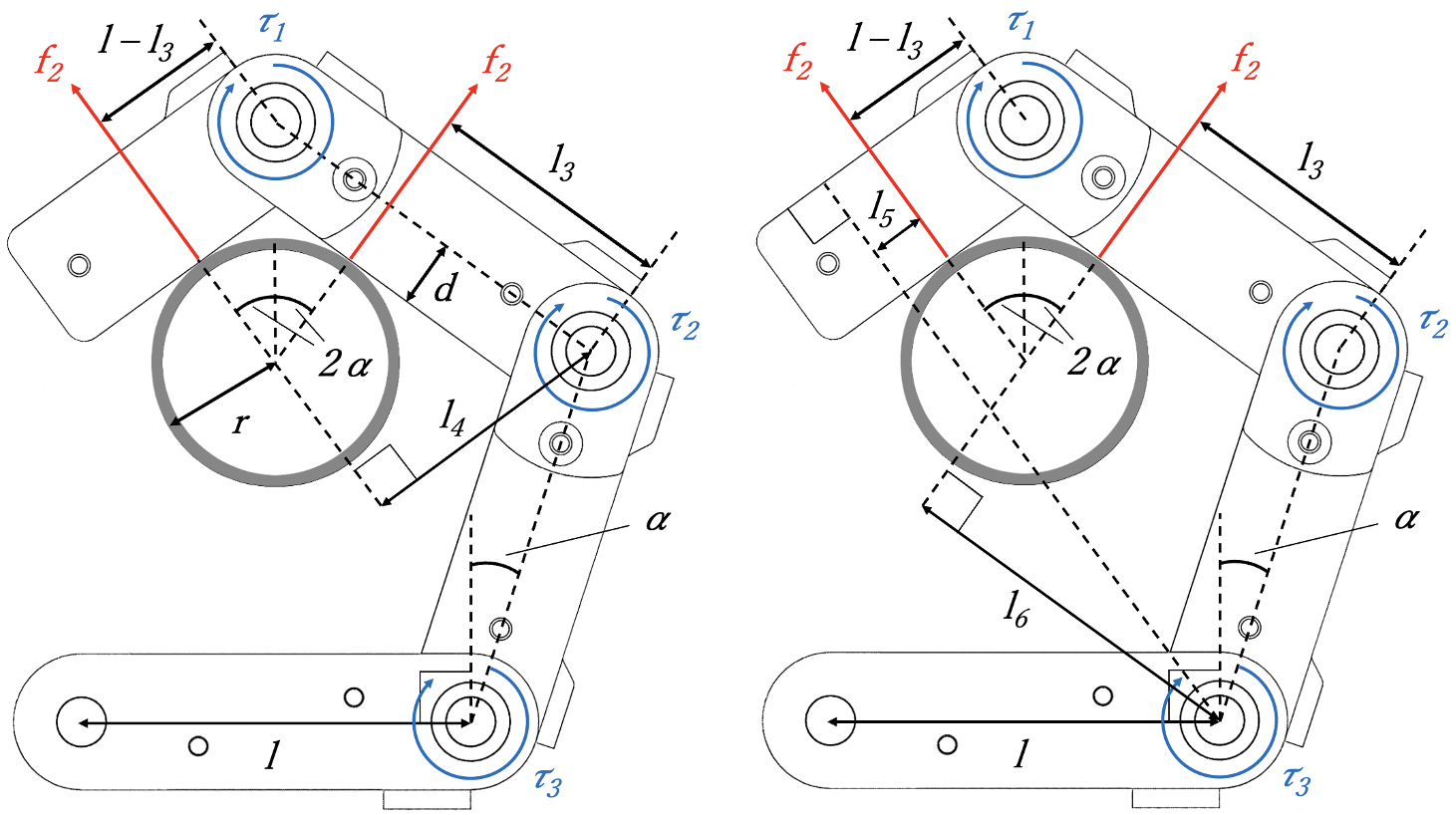}
        \caption{}
        \label{morealpha}
    \end{subfigure}
    \captionsetup{justification=raggedright, singlelinecheck=false}
\caption{\quad \textbf{Statics model when grasping object.}\, $R_1$ represents the radius of the built-in finger joint pulley. \textbf{(a)} Opening angle from 0 to less than $\pi / 10$. \textbf{(b)} Opening angle $\pi / 10$.}
    \vspace{-3mm}
\end{figure}

The prime motivation in our work is to use this hand for perching. Therefore, it is important to know how much load the hand can hold when performing dead-hang. Assuming that the fingers bear the weight evenly distributed, in the hand shape condition represented in Fig. \ref{lessalpha} the force $f_1$ acting on the finger and moment arm $l_1, l_2$ can be expressed geometrically using load m and opening angle $\alpha$:
\begin{equation}
    f_1 = \frac{mg}{3\cos(2\alpha)},\,
    \label{eq:f1}
\end{equation}
\begin{equation}
    l_1 = \frac{l}{2\cos(2\alpha)} + \frac{l\sin(\alpha)}{\cos(2\alpha)} - (r + d)\tan(2\alpha), \\
    l_2 = l_1 + l\sin(\alpha).
    \label{eq:l1}
\end{equation}

Besides, $f_1$ and servo stall torque $\tau_{st}$ along with the radius of the gear pulley $R_{gp}$ and gear reduction ratio $c$ satisfy the following tension equilibrium equation with (\ref{eq:euler}):
\begin{equation}
    3\,\frac{f_1l_1 + f_1l_2}{R_1} \leq c\frac{\tau_{st}}{R_{gp}}\,e^{\mu (2\pi + \alpha)}.
    \label{eq:calcm}
\end{equation}
When $\alpha$ is small enough, the second and third terms of $l_1$ converge to zero, therefore we assume that only the first term of $l_1$ is considered and the impact of the torque at the third joint is negligible. Based on this assumption and substituting equation (\ref{eq:f1}), (\ref{eq:l1}) into equation (\ref{eq:calcm}), maximum mass $m_{\text{max(a)}}$ can be expressed as follows:
\begin{equation}
    m_{\text{max(a)}} = \frac{\tau_{st}\,R_1\,c\,e^{\mu (2\pi + \alpha)}}{gl\,R_{gp}} \cdot \frac{\cos^2(2\alpha)}{\sin(\alpha)\cos(2\alpha) + 1}.
    \label{maxa}
\end{equation}
This value decreases monotonically as $\alpha$ increases. On the other hand, when $\alpha$ reaches $\frac{\pi}{10}$, the system passes to the state shown in Fig. \ref{morealpha}, where the load is supported at two points on each finger. Then, $m_{max(b)}$ can be expressed as follows:
\begin{equation}
    f_2 = \frac{mg}{6cos(2\alpha)},\, 
\end{equation}
\begin{equation}
    m_{\text{max(b)}} = \frac{\tau_{st}\,R_1\,c\,e^{\mu \frac{21\pi}{10}}}{gl\,R_{gp}} \cdot \frac{4\cos^2(2\alpha)}{4\sin(\alpha)\cos(2\alpha) + 5\cos(2\alpha) + 1}.
    \label{maxb}
\end{equation}


The increase in the contact points between the grasped object and the fingers from one to two results in a higher value of the maximum mass that the hand can support at $\alpha = \frac{\pi}{10}$. This jump in $m_{\text{max}}$ suggests the presence of a singularity point where the hand is expected to achieve a greater load bearing capacity as it opens up with increasing load, shown in Fig. \ref{morealpha}. This will be evaluated later in Section \ref{sec:expandres}.

The parameters $R_1$, $l$, $\tau_{st}$, and $R_{gp}$, which are determined during the actual fabrication of the hand, must be chosen so that $m_{\text{max(b)}}$ exceeds the load applied to the fingers during perching.

\section{C\footnotesize ONTROL}
\label{sec:control}

\subsection{Flight Control of Base Quadrotor}

\begin{figure}[tbp]
    \centering
    \includegraphics[keepaspectratio, width=0.6\linewidth]{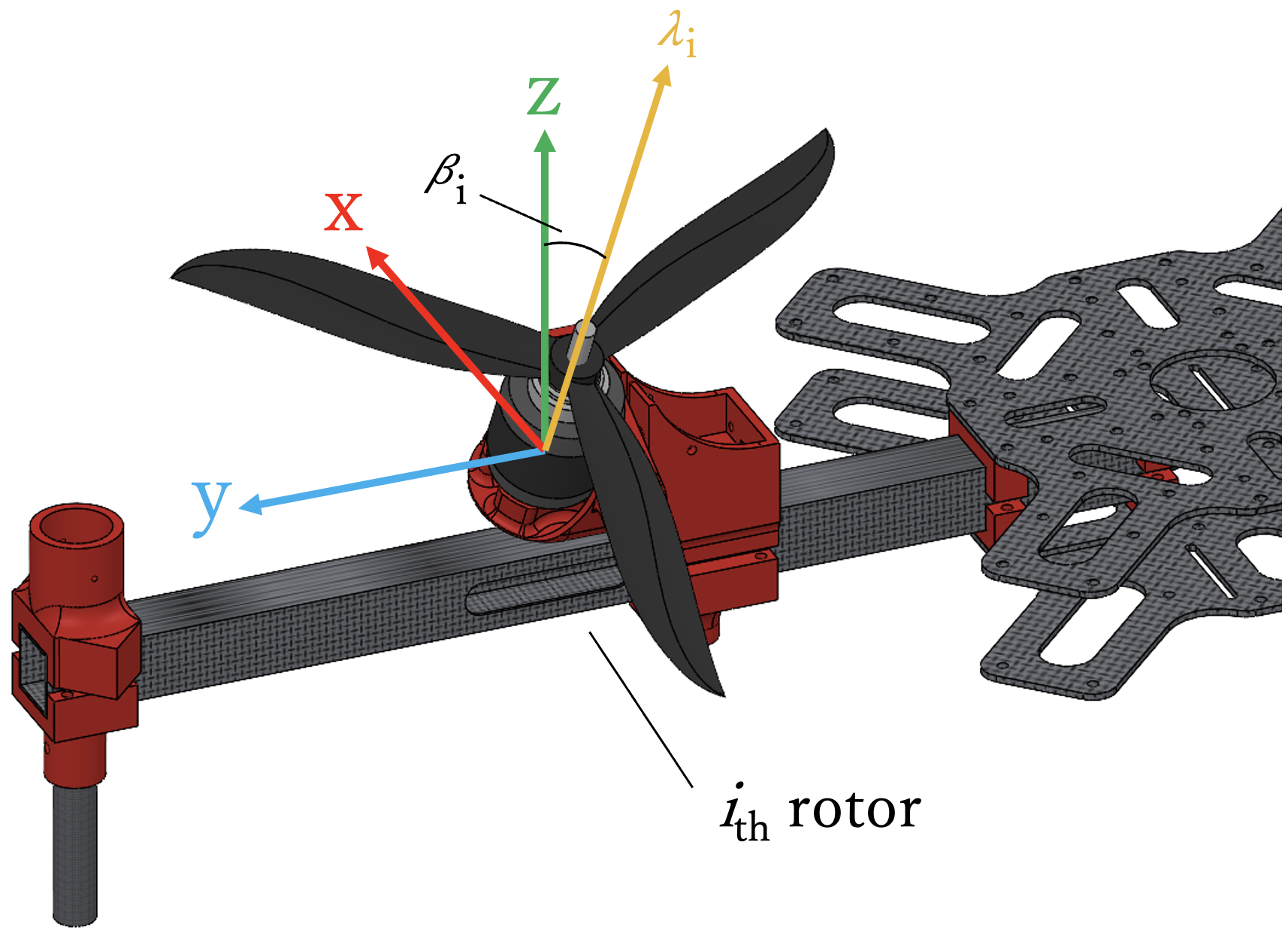}
    \captionsetup{justification=raggedright, singlelinecheck=false}
    \caption{\quad \textbf{Coordinate arrangement on the base quadrotor.}}
    \label{coordinate}
    \vspace{-3mm}
\end{figure}

This section explains the control and dynamics model of the fully actuated quadrotor\cite{j-sugihara} used as the base robot. First, as shown in Fig. \ref{coordinate}, each rotor module has its own coordinate system ${\{{U_i}\}}$, and the $i$-th thrust direction $\lambda_i$ in the $xz$-plane of ${\{{U_i}\}}$ and three-dimensional vector $\Lambda_i$ in the CoG coordinate ${\{{C}\}}$ can be defined as follows:
\begin{equation}
    \bm{\lambda}_i = 
    \begin{bmatrix}
        \lambda_{i,x} & \lambda_{i,z} \\
    \end{bmatrix}
    ^T,\\
    ^{\{{C}\}}\bm{\Lambda_i} = ^{\{{C}\}}\bm{U}_{\{{U_i}\}}\,\bm{B\,\lambda_i} = \bm{U'_i\,\lambda_i},
\end{equation}
where $^{\{{C}\}}\bm{U}_{\{{U_i}\}}$ is the rotation matrix from ${\{{U_i}\}}$ to ${\{{C}\}}$ and
\begin{equation}
    \bm{B} = 
    \begin{pmatrix}
        1 & 0 & 0 \\
        0 & 0 & 1 \\
    \end{pmatrix}
    ^T.
\end{equation}

Then, define the matrix $^{\{{C}\}}\Lambda$ which arranges the thrust vectors of each rotor:
\begin{equation}
    ^{\{{C}\}}\bm{\Lambda} = 
    \begin{bmatrix}
        {\bm{\Lambda}_1}^T & {\bm{\Lambda}_2}^T & {\bm{\Lambda}_3}^T & {\bm{\Lambda}_4}^T \\
    \end{bmatrix}
    ^T.
\end{equation}
With this thrust vector $^{\{{C}\}}\bm{\Lambda}$, the wrench matrix ${^{\{{C}\}}\bm{W}}$ which consists of force $^{\{{C}\}}\bm{f}$ and torque $^{\{{C}\}}\bm{\tau}$ can be expressed by following:
\begin{equation}
    ^{\{{C}\}}\bm{W} =
    \begin{bmatrix}
        ^{\{{C}\}}\bm{f} & ^{\{{C}\}}\bm{\tau} 
    \end{bmatrix}
    ^T
    = \bm{Q}\,^{\{{C}\}}\bm{\Lambda},
\end{equation}
where
\begin{equation}
    \bm{Q} =
    \begin{pmatrix}
         \bm{E^{3 \times 3}}\,\bm{E^{3 \times 3}}\,\bm{E^{3 \times 3}}\,\bm{E^{3 \times 3}} \\
         \begin{bmatrix}
             ^{\{{C}\}}\bm{p_i} \times
         \end{bmatrix}
         - \kappa\,\sigma_i\,\bm{E^{3 \times 3}}
    \end{pmatrix}
    ^T,
\end{equation}
with $i$-th rotor position $p_i$, rotor counter torque coefficient $\kappa$ and $i$-th rotor's rotation direction $\sigma_i$ that take a value of either -1 or 1. 

Thereby, thrust matrix $^{\{{C}\}}\bm{\lambda}^{des}$ can be calculated as follows:
\begin{equation}
    \bm{\lambda}^{des} = (\,\bm{Q}\,\text{diag}(U'_1 \cdots U'_4)\,)^\#\, \bm{W}^{des},\\
\end{equation}
with the pseudo-inverse matrix $(\,\bm{Q}\,diag(U'_1 \cdots U'_4)\,)^\#$ and the desired wrench $\bm{W}^{des}$. Then the scalar thrust force $\lambda_i^{out}$ and rotor vector angle $\beta_i$ for the $i$-th rotor can be expressed as follows:
\begin{equation}
    \lambda_i^{out} = \| \bm{\lambda}_i \|,\,
    \beta_i = \text{atan2}
    \begin{pmatrix}
        {\lambda}_{i,z}, & {\lambda}_{i,x}
    \end{pmatrix}.
\end{equation}

For calculating $\bm{W}^{des}$, calculate $\bm{f}^{des}_{PID}$ and $\bm{\tau}^{des}_{PID}$ using PID control.

\subsection{Motion Planning for Pendulum Perching}

When pendulum perching is performed, the robot body rests in an almost vertical position; hence, the rate of change in position and rotation are used to determine landing. When the velocity and angular velocity fall below a certain threshold while in the landing state, it is determined as the landing trigger, and the rotor thrust is cut off. Using halt causes the pendulum motion after perching to become larger, while it allows for quicker perching.

During detachment, the thrust force gradually increases and the robot body returns to a horizontal state from vertical state. However, since the hand is constrained by the crossbeam during the detachment maneuver, only the z position and the pitch angle change. Hence, slight discrepancies can be amplified by the integral term of PID control during takeoff and potentially leading to unstable flight behavior after detachment. To mitigate this issue, we compulsorily set integral terms 0 except z during takeoff and start recalculating integral terms after detachment to ensure smooth flight. 

The target position for detachment is set as follows using the CoG position $\bm{r}^{bottom}$ at the lowest point:
\begin{equation}
    \bm{r}^{des} = \bm{r}^{bottom} -
    \begin{bmatrix}
        l_h(\cos(\theta_p^{des}) - cos(\theta_p))\cos(\theta_y) \\
        l_h(\cos(\theta_p^{des}) - cos(\theta_p))\sin(\theta_y) \\
        -l_h(\sin(\theta_p^{des}) - sin(\theta_p))
        \label{eq:targetposset}
    \end{bmatrix},
\end{equation}
where $l_h$ is the distance between the crossbeam and CoG, $\theta_p, \theta_y$ is the pitch and yaw angle of CoG. The validity of this target calculation is demonstrated in Section \ref{sec:expandres}.

\begin{figure}[t]
    \centering
    \includegraphics[keepaspectratio, width=0.8\linewidth]{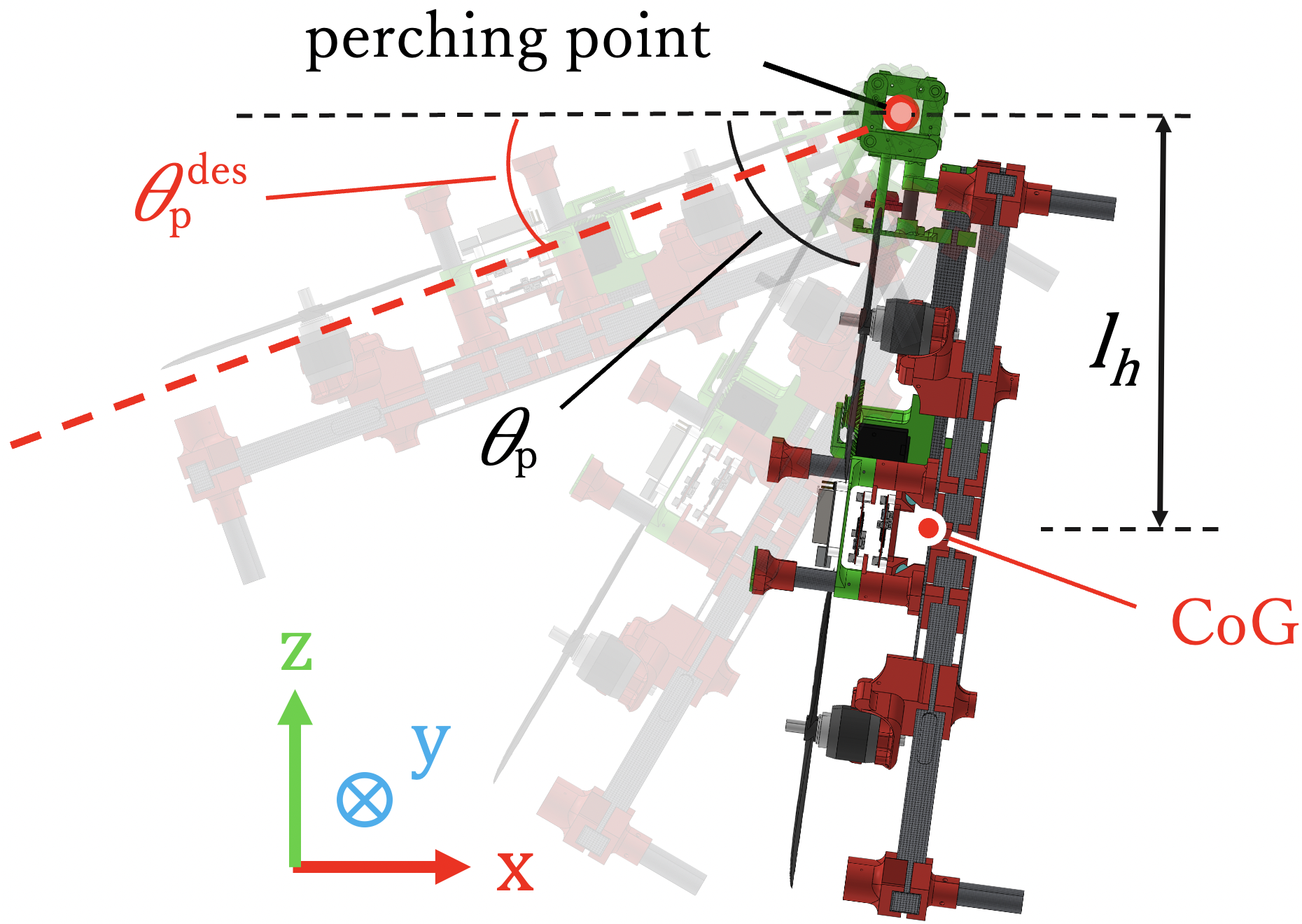}
    \hspace{1mm}
    \captionsetup{justification=raggedright, singlelinecheck=false}
    \caption{\quad \textbf{Diagram for target position settings during detachment.}}
    \label{rotatediagram}
    \vspace{-3mm}
\end{figure}


\section{E\footnotesize XPERIMENTS \normalsize \& R\footnotesize ESULTS}
\label{sec:expandres}

\subsection{Hand Prototype}

In this research, we used PLA mainly for the hand because of their lightweightness, including the bushings in the finger joints and the spherical joint where relatively strong forces are applied. Hollow CFRP cylinder pipes and columns are used as the arm beams and parts of the hand's support structure. Dynamixel XH350-W210R used as the actuator doubles the torque using a pair of double helical gears. Almost all tendons are fixed by setscrews, except for the active tendons fixed to the pulley on the larger helical gear by stainless steel sleeves. There were three planes inside the fingers through which the tendons pass; the two outer planes carried passive tendons, while the central plane carried active tendons. 

The size of the hand unit including the servo motor was approximately 243 mm $\times$ 57 mm $\times$ 260 mm, and the total weight was 390 g.

\begin{figure}[h]
    \centering
    \includegraphics[keepaspectratio, width=\linewidth]{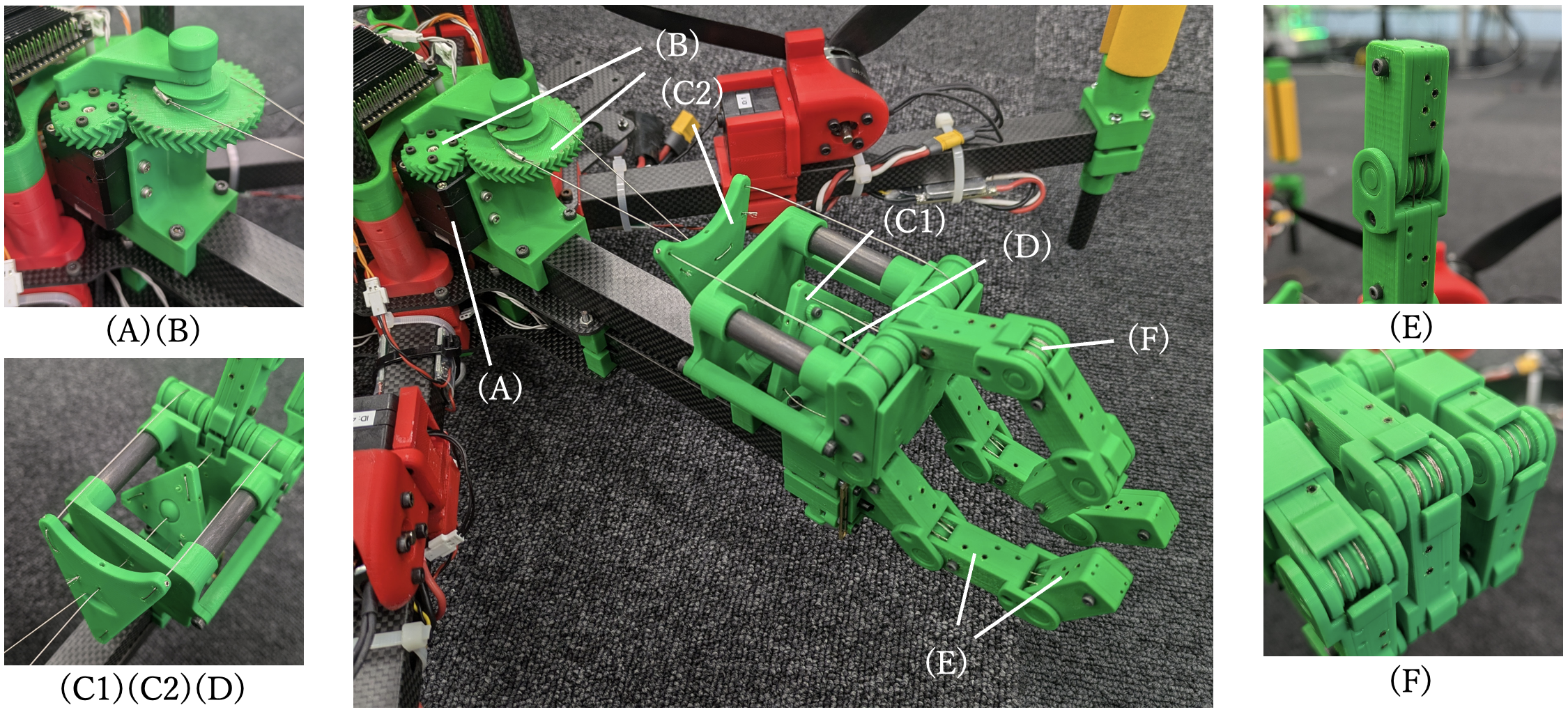}
    \caption{\quad \textbf{Tri-force hand prototype.} \textbf{(A)} Servo motor as actuator. \textbf{(B)} Gear reduction mechanism. \textbf{(C1)(C2)} Triangular two-dimensional differentials. \textbf{(D)} Spherical joint. \textbf{(E)} Wire fixation point. \textbf{(F)} Finger joint and built-in joint pulley with three planes for passive / active tendons.}
    \vspace{-3mm}
\end{figure}

\subsection{Basic Experimental Evaluation}

\subsubsection{Hand Adaptability Test}

\begin{figure}[tbp]
    \centering
    \begin{subfigure}[tb]{0.24\linewidth}
    \centering
        \includegraphics[keepaspectratio, width=\linewidth]{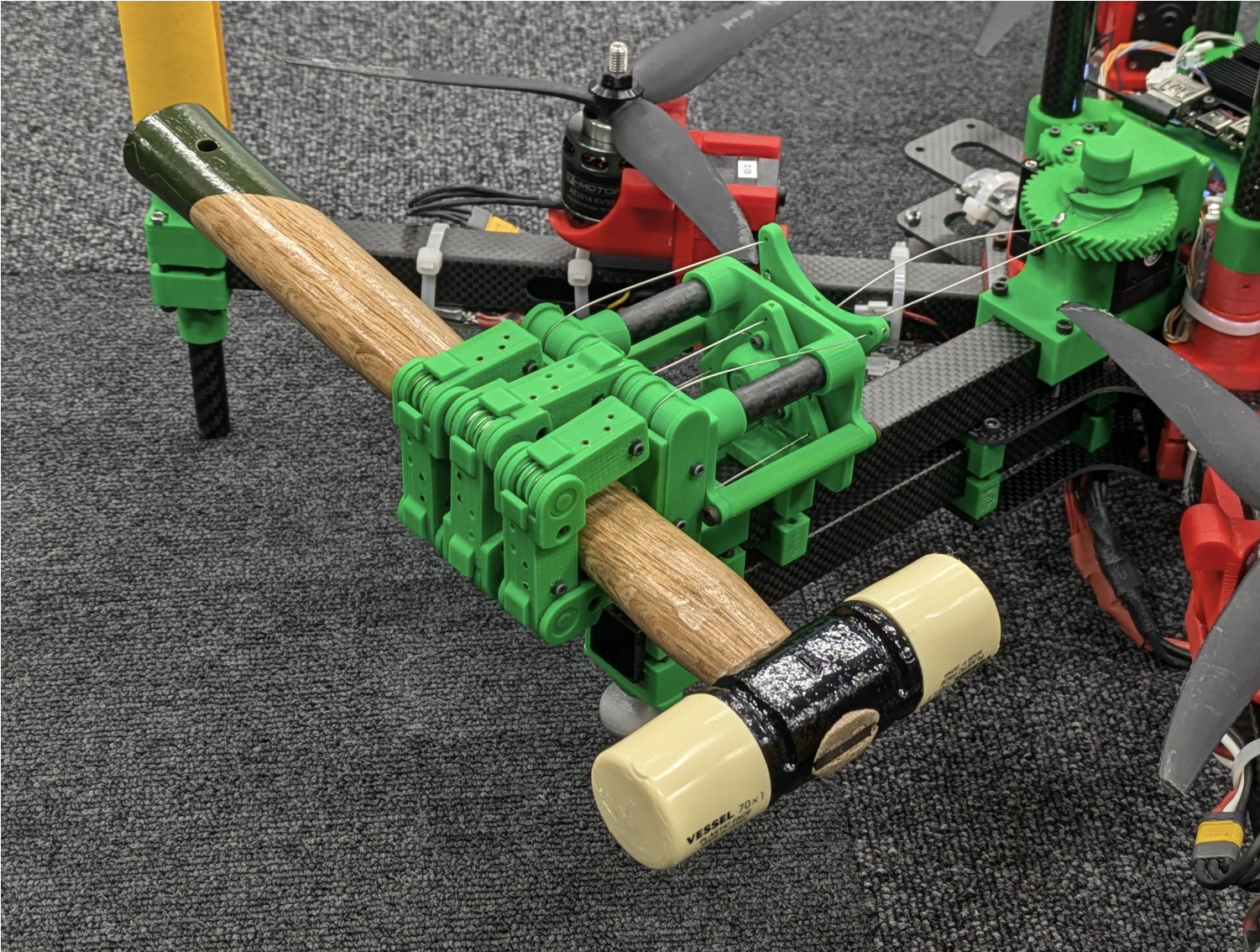}
        \caption{}
        \vspace{-1mm}
        \label{hammer}
        \hspace{1mm}
    \end{subfigure}
    \hfill
    \begin{subfigure}[tb]{0.24\linewidth}
        \centering
        \includegraphics[keepaspectratio, width=\linewidth]{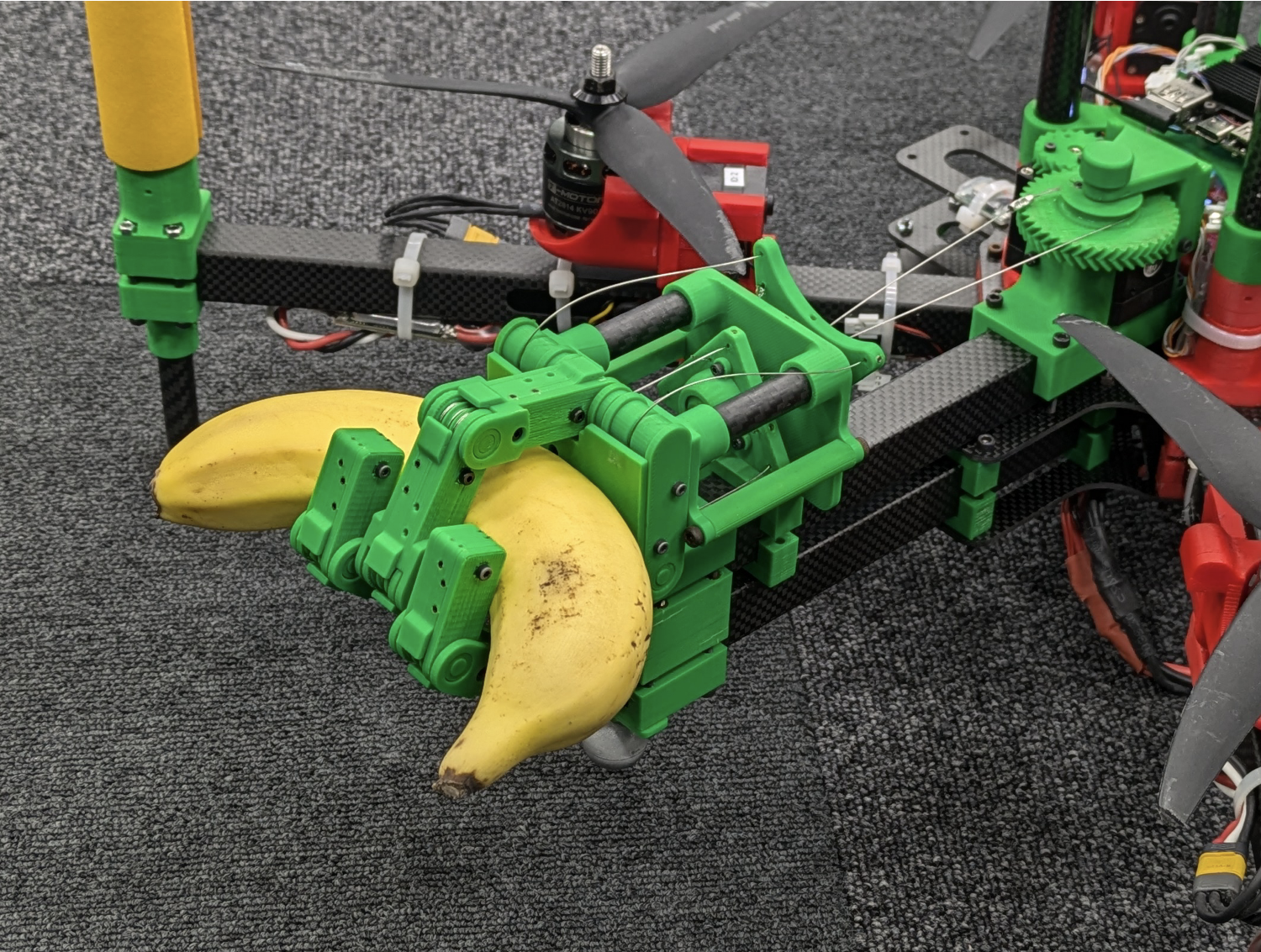}
        \caption{}
        \label{banana1}
        \vspace{-1mm}
        \hspace{1mm}
    \end{subfigure}
    \hfill
    \begin{subfigure}[tb]{0.24\linewidth}
    \centering
        \includegraphics[keepaspectratio, width=\linewidth]{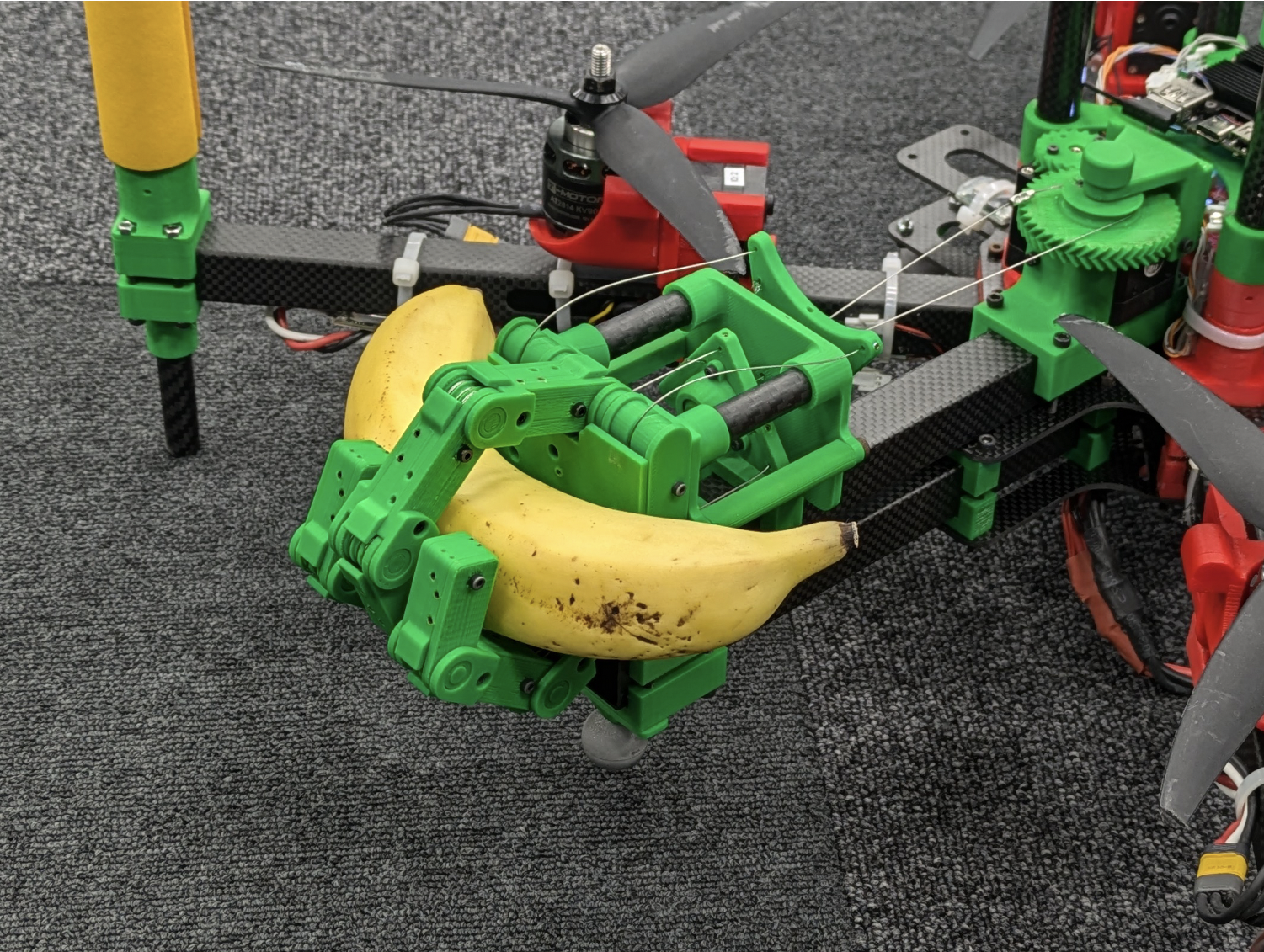}
        \caption{}
        \label{banana2}
        \vspace{-1mm}
        \hspace{1mm}
    \end{subfigure}
    \hfill
    \begin{subfigure}[tb]{0.24\linewidth}
    \centering
        \includegraphics[keepaspectratio, width=\linewidth]{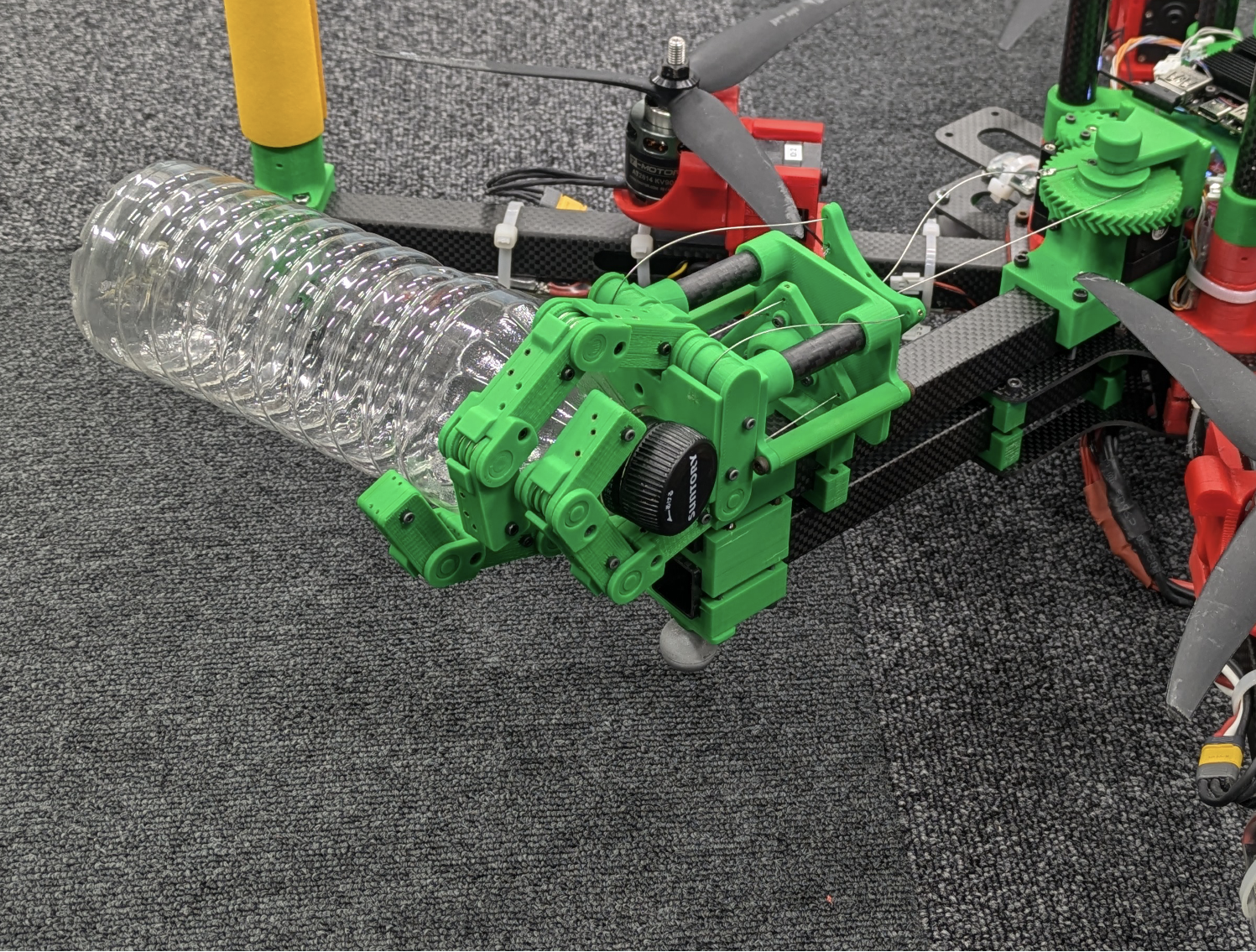}
        \caption{}
        \vspace{-1mm}
        \label{bottle}
        \hspace{1mm}
    \end{subfigure}
    \begin{subfigure}[tb]{0.24\linewidth}
        \centering
        \includegraphics[keepaspectratio, width=\linewidth]{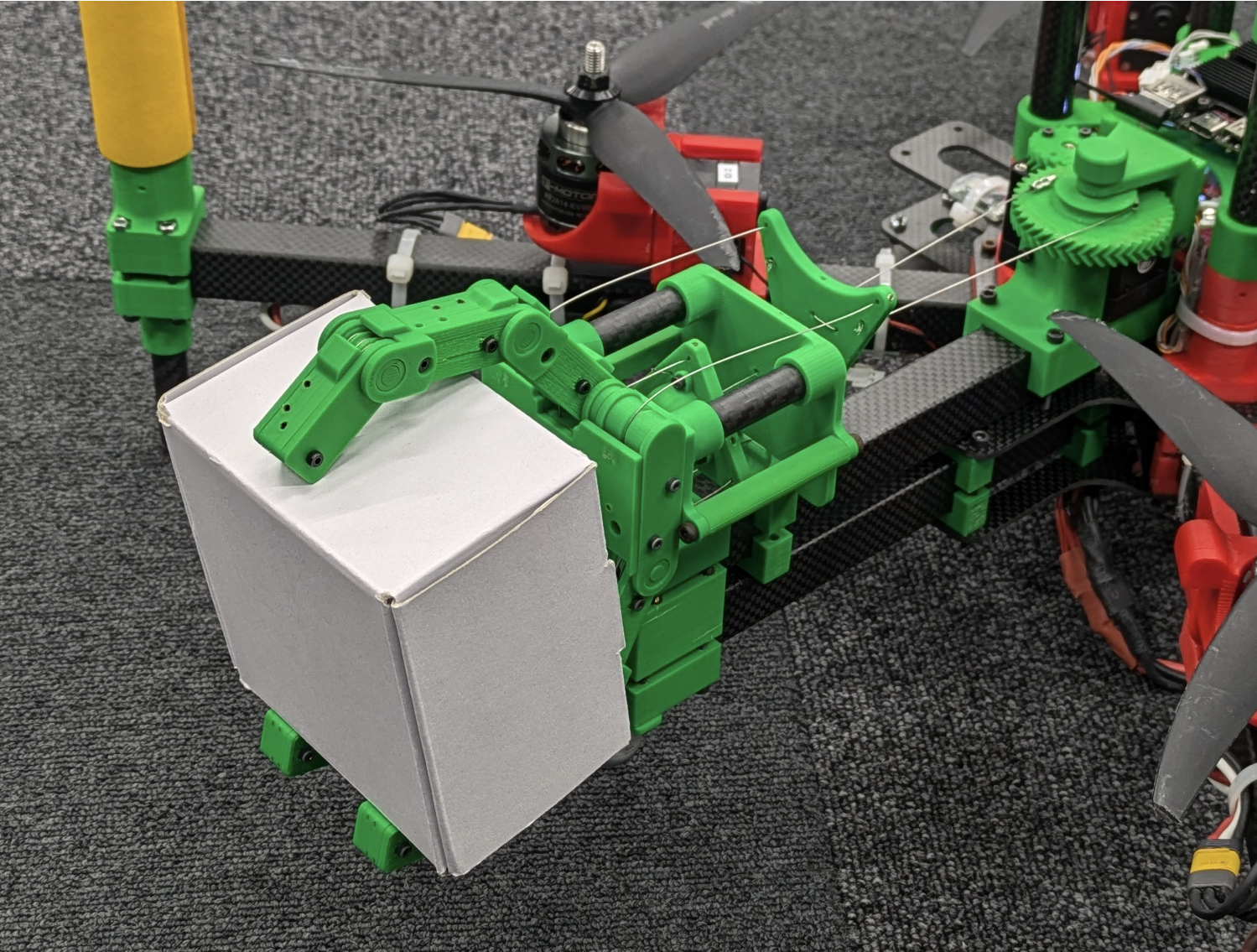}
        \caption{}
        \label{box}
        \vspace{-3mm}
        \hspace{1mm}
    \end{subfigure}
    \hfill
    \begin{subfigure}[tb]{0.24\linewidth}
    \centering
        \includegraphics[keepaspectratio, width=\linewidth]{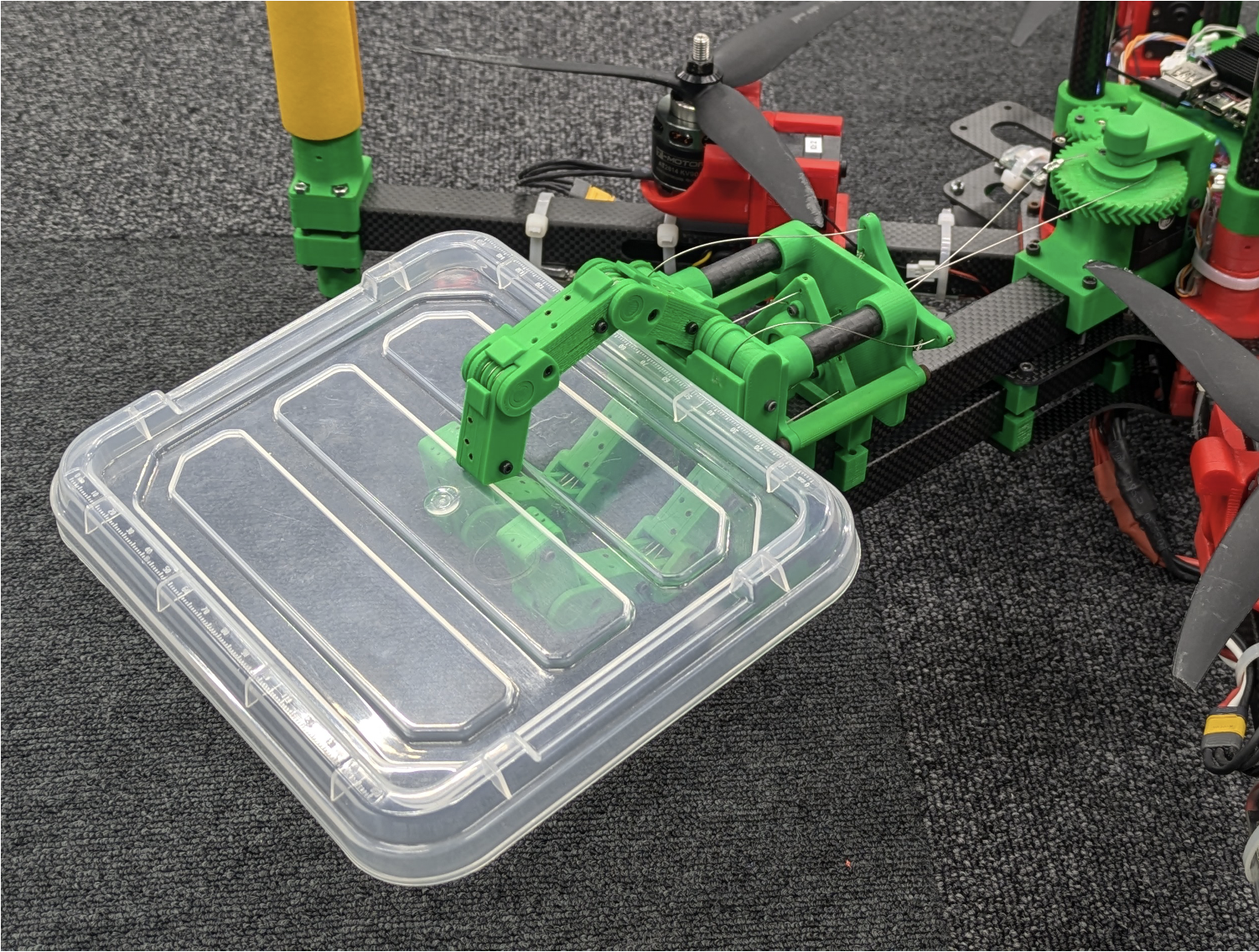}
        \caption{}
        \label{lid}
        \vspace{-3mm}
        \hspace{1mm}
    \end{subfigure}
    \hfill
    \begin{subfigure}[tb]{0.24\linewidth}
    \centering
        \includegraphics[keepaspectratio, width=\linewidth]{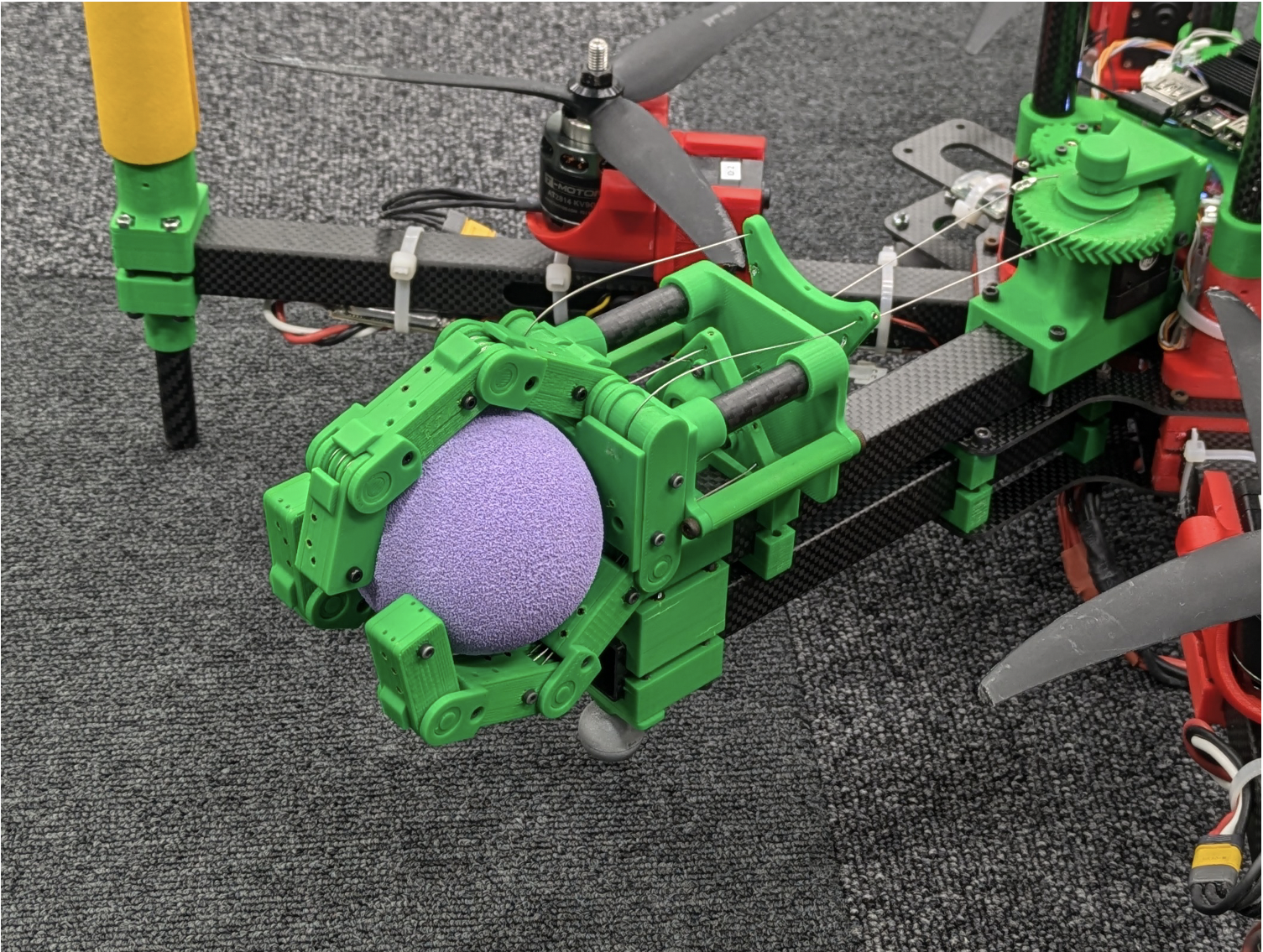}
        \caption{}
        \label{ball}
        \vspace{-3mm}
        \hspace{1mm}
    \end{subfigure}
    \hfill
    \begin{subfigure}[tb]{0.24\linewidth}
    \centering
        \includegraphics[keepaspectratio, width=\linewidth]{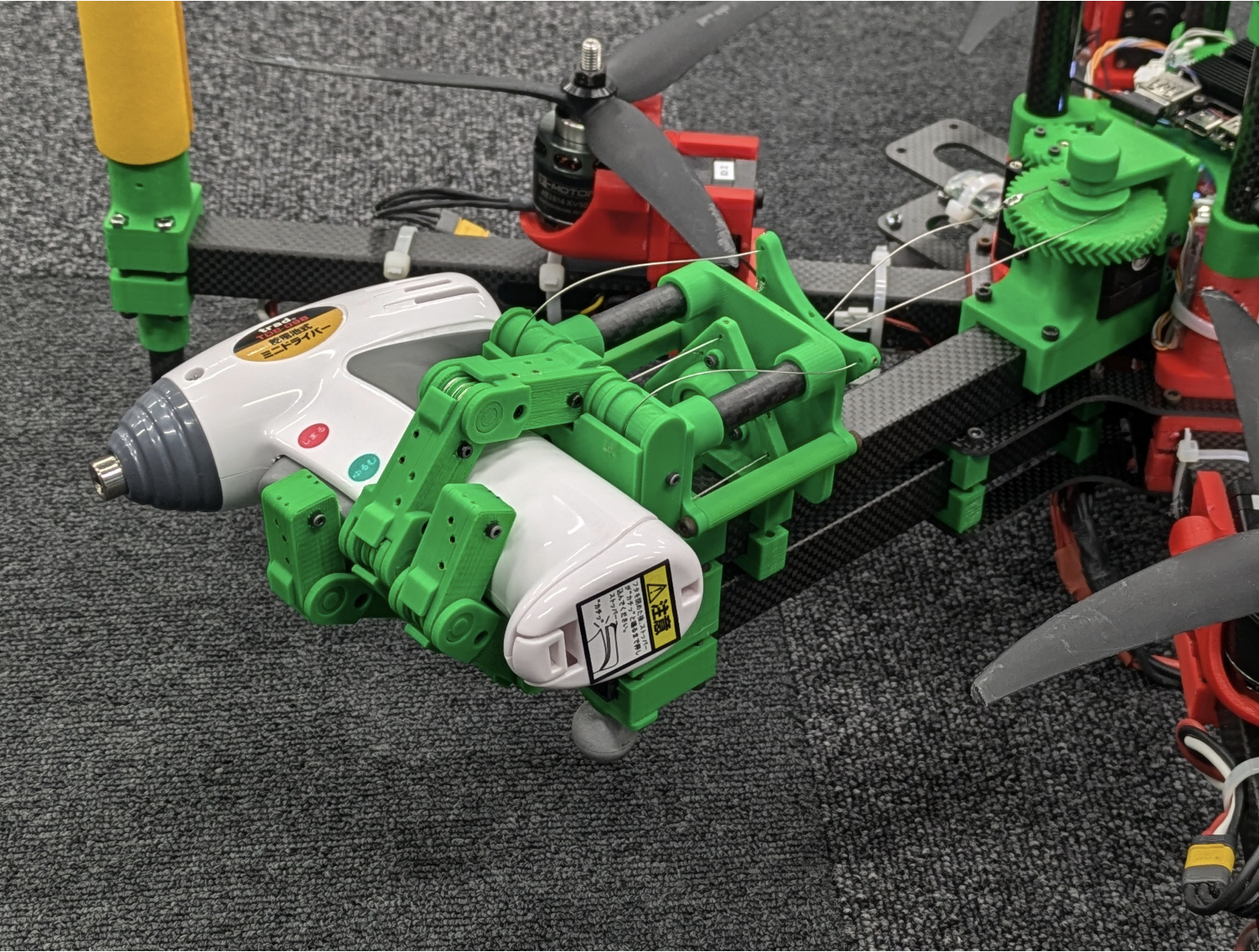}
        \caption{}
        \label{drill}
        \vspace{-3mm}
        \hspace{1mm}
    \end{subfigure}
    \captionsetup{justification=raggedright, singlelinecheck=false}
    \caption{\quad \textbf{Tri-force Hand Grasping objects with various shape.}}
    \vspace{-3mm}
\end{figure}

To demonstrate the adaptability of the hand, we performed experiments grasping objects of various shapes. In addition to the common-shaped object shown in Fig. \ref{hammer}, the hand was capable of grasping objects with a concave shape in center that conventional straight differential cannot grasp precisely shown in Fig. \ref{banana1}, \ref{banana2}, and other types of object such as bottle neck with significantly different diameters, box, plate, ball, and actual power drill shown in Fig. \ref{bottle}--\ref{drill}.

\subsubsection{Hand Load Bearing Capacity Test}

Since the aerial robot used in this study performs multi-directional perching and generates centrifugal or centripetal forces during perching sequence, the hand must withstand the load other than its own weight. Therefore, we evaluated the grip strength of the hand with dead-hang motion.

\begin{figure}[tbp]
    \centering
    \begin{subfigure}[tb]{0.3\linewidth}
        \centering
        \includegraphics[keepaspectratio, width=\linewidth]{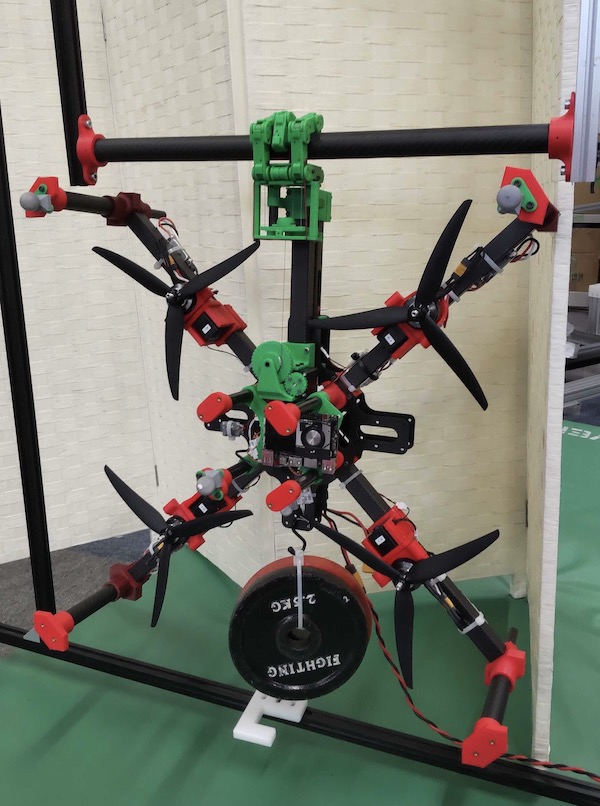}
        \caption{}
        \label{testenv}
        \vspace{-2mm}
        \hspace{1mm}
    \end{subfigure}
    \hfill
    \begin{subfigure}[tb]{0.68\linewidth}
    \centering
        \includegraphics[keepaspectratio, width=\linewidth]{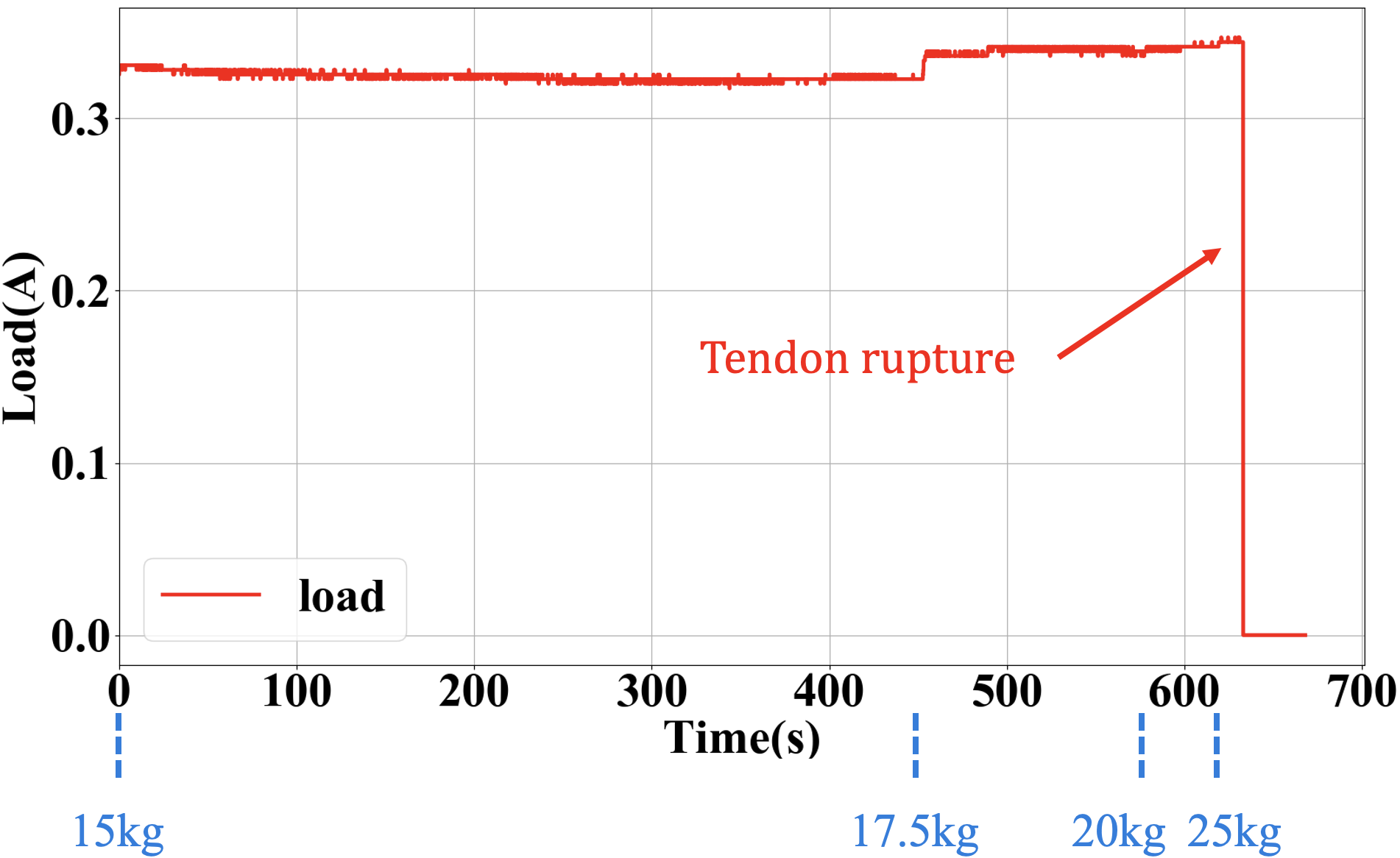}
        \caption{}
        \label{loadplot}
        \vspace{-2mm}
        \hspace{1mm}
    \end{subfigure}
    \captionsetup{justification=raggedright, singlelinecheck=false}
    \caption{\quad \textbf{Hand load bearing capacity test.} \textbf{(a)} Experiment environment setup. \textbf{(b)} Plots of servo load. The servo motor load is represented in the form of an AD-converted current value.}
\end{figure}

As shown in Fig. \ref{testenv}, set the CFRP pipe cross beam with a diameter of 24 mm, which represents the maximum diameter that can be held by the hand with all fingers fully closed. A weight was applied to the base of the robot body in increments of 2.5 kg, and the load at the point when the hand was detached from the pipe or broken was defined as the load bearing capacity. Applying the parameters to equations (\ref{maxa}) and (\ref{maxb}), $m_{\text{max(a)}}$ and $m_{\text{max(b)}}$ were calculated as $23.9$ kg and $29.6$ kg on this hand.

As a result, the hand supported 27.5 kg, including 2.5 kg of its own weight until the robot body fell due to rupture of the active tendon in the spherical joint. Fig. \ref{loadplot} shows the graph of the servo load from the no-load condition to the point of fall, where the servo load symbolized by the red line drops significantly. 
From the plot, it can be observed that the servo load remained almost unchanged throughout except for a little increase around 450 seconds. This time corresponds to the point when 17.5 kg was added, and it was observed that the first joints of the fingers aligned as shown in Fig. \ref{morealpha}. This phenomenon was considered to be attributed to the abrupt change in load bearing capacity due to the degree of hand opening described in Section \ref{sec:design}. It indicated that the servo load was supported by the geometric structure of the hand when the opening angle of the finger reached the singularity point due to an increase in weight, resulting in the transition from the state shown in Fig. \ref{lessalpha} to that in Fig. \ref{morealpha}. This result demonstrated the significantly large grip force of this hand considering its lightweightness.


\begin{figure}[tbp]
    \begin{subfigure}[tb]{0.49\linewidth}
    \centering
        \includegraphics[keepaspectratio, width=\linewidth]{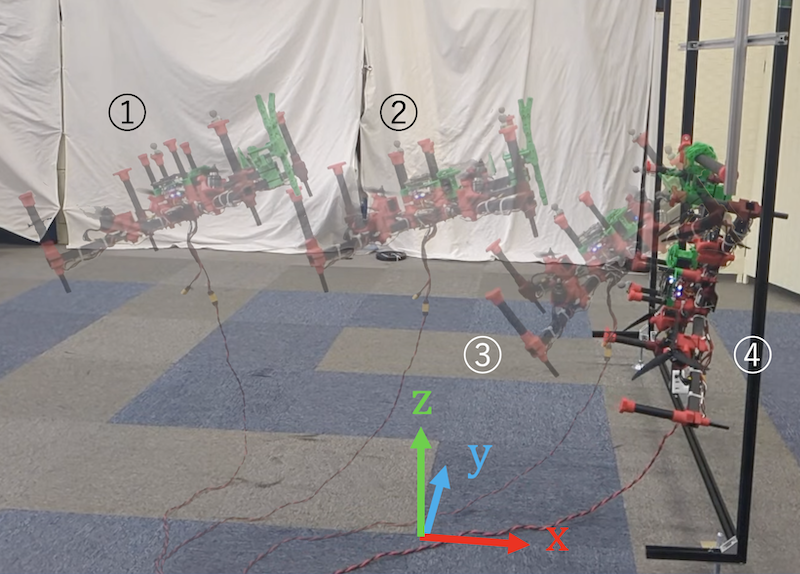}
        \caption{Attachment}
        \label{attachfig}
        \vspace{-2mm}
        \hspace{1mm}
    \end{subfigure}
    \begin{subfigure}[tb]{0.49\linewidth}
    \centering
        \includegraphics[keepaspectratio, width=\linewidth]{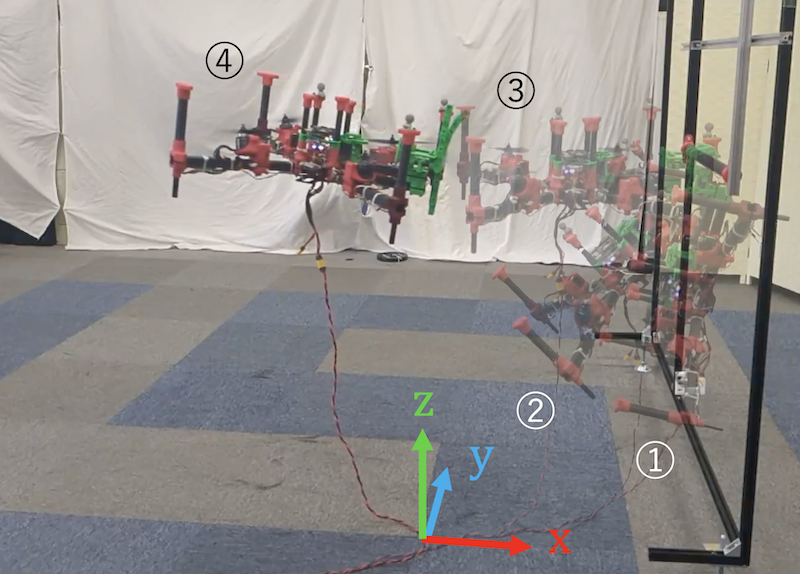}
        \caption{Detachment}
        \label{detachfig}
        \vspace{-2mm}
        \hspace{1mm}
    \end{subfigure}
    \captionsetup{justification=raggedright, singlelinecheck=false}
    \caption{\quad \textbf{Perching experiment setup.}  \textbf{(a)} Trajectory of the quadrotor performing oblique perching. \textbf{(b)} Trajectory of the takeoff from vertical state and detachment.}
    \vspace{-3mm}
\end{figure}

\subsubsection{Perching Test}

To demonstrate the capability of multi-directional perching, we used the experimental setup shown in Fig. \ref{testenv} and performed oblique perching experiment using a CFRP pipe with a diameter of 24 mm and a metal square column with a side length of 20 mm. Fig. \ref{attachfig} and Fig. \ref{detachfig} shows the sequence of the perching and detachment experiment to the cylinder pipe. Takeoff operation is performed according to the control method described in Section \ref{sec:control}.

\begin{figure}[tbp]
    \centering
    \begin{subfigure}[tb]{0.49\linewidth}
        \includegraphics[keepaspectratio, width=\linewidth]{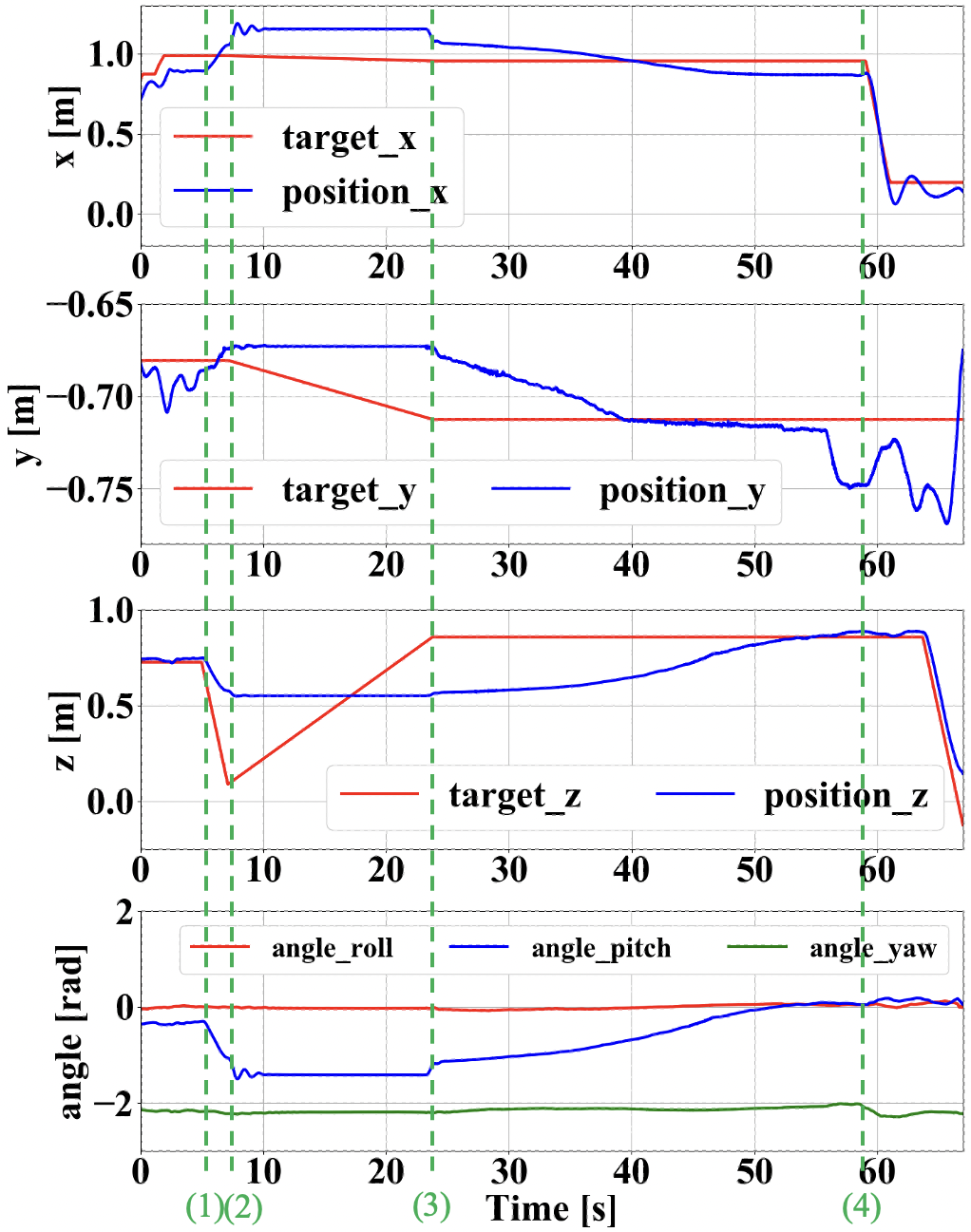}
        \caption{Cylinder pipe}
        \label{normalplot}
    \end{subfigure}
    \centering
    \begin{subfigure}[tb]{0.49\linewidth}
        \includegraphics[keepaspectratio, width=\linewidth]{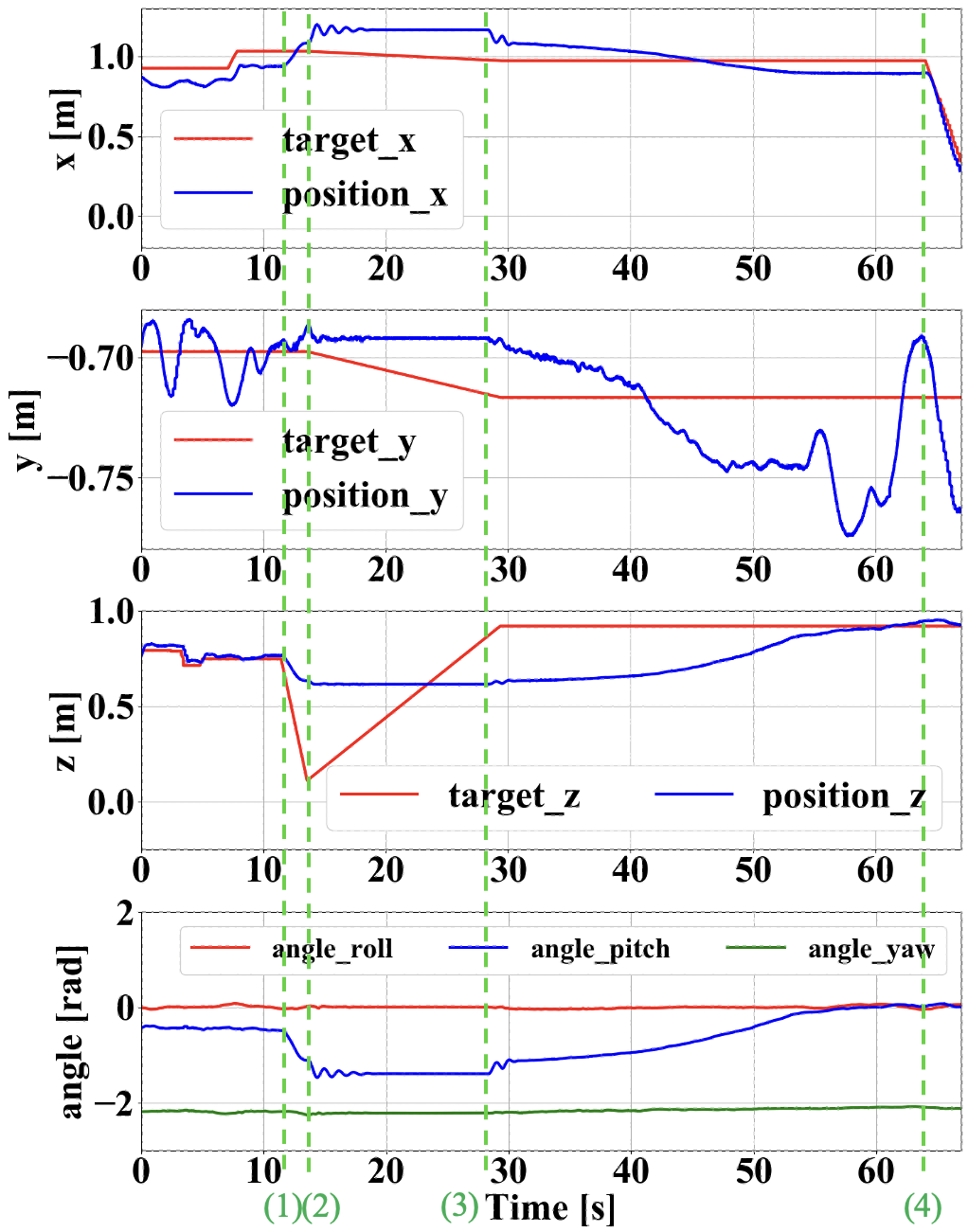}
        \caption{Square column}
        \label{squareplot}
    \end{subfigure}
    \caption{\quad \textbf{Plots of current/target position and angle during pendulum perching sequence.}  Attachment at (1), landing at (2), takeoff from vertical state at (3) and detachment at (4).}
\end{figure}

Fig. \ref{normalplot} shows the error plot between the current position and the target position during experiment. From the plot, it can be observed that the position x, y, and z correctly converges to target value, and the unstable oscillations in the x-direction seen immediately after detachment are quickly reduced due to the recalculation of the integral term. Furthermore, the smooth convergence of the pitch angle towards 0 indicates that the target position setting shown in (\ref{eq:targetposset}) was appropriate. The oscillation observed in the y-direction is attributed to the fact that the fingers do not open simultaneously, leading to slight positional deviations during detachment.

\begin{figure}[tbp]
    \centering
    \includegraphics[keepaspectratio, width=0.8\linewidth]{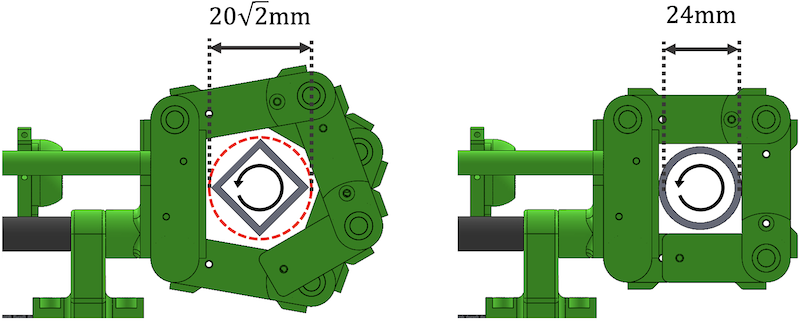}
    \captionsetup{justification=raggedright, singlelinecheck=false}
    \caption{\quad \textbf{Perching to square column/cylinder pipe.} To perform perching for the square column, adjusting finger angle is indispensable due to the huge servo motor load.}
    \label{columngrip}
    \vspace{-3mm}
\end{figure}

Perching experiment on a square column was also conducted. However, while the diameter of the cylinder remains constant, the diameter of a square column changes with the rotation of the robot body during perching as shown in Fig. \ref{rotatediagram}, reaching a maximum value when the pitch angle reaches 45 degree. Therefore, when perching on the square column while gripping it as firmly as the cylinder pipe, a significant force acts in the direction of opening the hand during \ding{174} to \ding{175} in Fig. \ref{attachfig} and \ding{172} to \ding{174} in Fig. \ref{detachfig}, which may result in the servo overload. To solve this issue, it is necessary to adjust the joint angle so that the square column can rotate freely without firmly interfering with the fingers. By adjusting the angle of the finger to match that used to grasp a cylinder pipe with a diameter of the circumscribed circle of the column as shown in Fig. \ref{columngrip}, we succeeded in perching and detachment without overloading the servo. Due to the small interference caused by the joints and edges of the column, the pitch angle plot shown in Fig. \ref{squareplot} exhibits a small oscillation around $t = 29$, which is not seen in Fig. \ref{normalplot}. The oscillation is sufficiently small, and the operation resumes smoothly afterwards. Nevertheless, the achievement of pendulum perching and detachment demonstrate the feasibility of our hand regarding grasping from multi-direction.



\section{C\footnotesize ONCLUSIONS}
\label{sec:conclusion}

This paper presents a hand mechanism named Tri-force hand, capable of both perching on and grasping objects adaptively. The hand employs tendon-driven actuation, making it lightweight and durable. Furthermore, by introducing the Tri-force system, two-dimensional differential mechanism, the hand achieves a three-fingered opposing configuration that can be adaptively and powerfully driven by a single actuator. When mounted on a fully actuated quadrotor, this system enables stable perching from multiple directions. Through perching and load bearing experiments, we demonstrated the advantages of the proposed hand mechanism.

There are some remaining challenges regarding the versatility of grasping. For example, the pinching action for a thin object such as a pen is still unstable with the current hand. In the future, we will add additional DoF for the finger and palm to achieve a more complex grasping motion. 
Furthermore, by adding rotation DoF to the wrist, the hand can perform not only grasping but also handling tools.
Regarding the application with this hand, agile perching and detaching will be investigated by developing the model predictive control. Meanwhile, aerial manipulation with the proposed hand grasping tools is also another crucial future work.  



\addtolength{\textheight}{0cm}   


\bibliographystyle{bibtex/IEEEtran}
\bibliography{bibtex/IEEEabrv, bibtex/references}

\begin{thebibliography}{10}
\providecommand{\url}[1]{#1}
\csname url@samestyle\endcsname
\providecommand{\newblock}{\relax}
\providecommand{\bibinfo}[2]{#2}
\providecommand{\BIBentrySTDinterwordspacing}{\spaceskip=0pt\relax}
\providecommand{\BIBentryALTinterwordstretchfactor}{4}
\providecommand{\BIBentryALTinterwordspacing}{\spaceskip=\fontdimen2\font plus
\BIBentryALTinterwordstretchfactor\fontdimen3\font minus \fontdimen4\font\relax}
\providecommand{\BIBforeignlanguage}[2]{{%
\expandafter\ifx\csname l@#1\endcsname\relax
\typeout{** WARNING: IEEEtran.bst: No hyphenation pattern has been}%
\typeout{** loaded for the language `#1'. Using the pattern for}%
\typeout{** the default language instead.}%
\else
\language=\csname l@#1\endcsname
\fi
#2}}
\providecommand{\BIBdecl}{\relax}
\BIBdecl

\bibitem{observation}
R.~Bonatti, Y.~Zhang, S.~Choudhury, W.~Wang, and S.~Scherer, ``Autonomous drone cinematographer: Using artistic principles to create smooth, safe, occlusion-free trajectories for aerial filming,'' in \emph{Proceedings of the 2018 International Symposium on Experimental Robotics}, Cham, 2020, pp. 119--129.

\bibitem{mapping}
\BIBentryALTinterwordspacing
J.~G. Monteiro, J.~L. Jiménez, F.~Gizzi, P.~Přikryl, J.~S. Lefcheck, R.~S. Santos, and J.~Canning-Clode, ``Novel approach to enhance coastal habitat and biotope mapping with drone aerial imagery analysis,'' \emph{Scientific Reports}, vol.~11, no.~1, p. 574, 2021.
\BIBentrySTDinterwordspacing

\bibitem{disaster}
N.~Michael, S.~Shen, K.~Mohta, V.~Kumar, K.~Nagatani, Y.~Okada, S.~Kiribayashi, K.~Otake, K.~Yoshida, K.~Ohno, E.~Takeuchi, and S.~Tadokoro, \emph{Collaborative Mapping of an Earthquake Damaged Building via Ground and Aerial Robots}.\hskip 1em plus 0.5em minus 0.4em\relax Berlin, Heidelberg: Springer Berlin Heidelberg, 2014, pp. 33--47.

\bibitem{underhandperching}
\BIBentryALTinterwordspacing
K.~Hang, X.~Lyu, H.~Song, J.~A. Stork, A.~M. Dollar, D.~Kragic, and F.~Zhang, ``Perching and resting—a paradigm for uav maneuvering with modularized landing gears,'' \emph{Science Robotics}, vol.~4, no.~28, p. eaau6637, 2019.
\BIBentrySTDinterwordspacing

\bibitem{underhandperching2}
\BIBentryALTinterwordspacing
S.~Ubellacker, A.~Ray, J.~M. Bern, J.~Strader, and L.~Carlone, ``High-speed aerial grasping using a soft drone with onboard perception,'' \emph{npj Robotics}, vol.~2, no.~1, p.~5, 2024.
\BIBentrySTDinterwordspacing

\bibitem{underhandperching3}
\BIBentryALTinterwordspacing
P.~M. Nadan, T.~M. Anthony, D.~M. Michael, J.~B. Pflueger, M.~S. Sethi, K.~N. Shimazu, M.~Tieu, and C.~L. Lee, ``{A Bird-Inspired Perching Landing Gear System1},'' \emph{Journal of Mechanisms and Robotics}, vol.~11, no.~6, p. 061002, 09 2019.
\BIBentrySTDinterwordspacing

\bibitem{underhandperching4}
\BIBentryALTinterwordspacing
F.~J.~G. Rubiales, P.~R. Soria, B.~C. Arrue, and A.~Ollero, ``Soft-tentacle gripper for pipe crawling to inspect industrial facilities using uavs,'' \emph{Sensors (Basel, Switzerland)}, vol.~21, 2021.
\BIBentrySTDinterwordspacing

\bibitem{overheadperching}
\BIBentryALTinterwordspacing
T.~Ching, J.~Z.~W. Lee, S.~K.~H. Win, L.~S.~T. Win, D.~Sufiyan, C.~P.~X. Lim, N.~Nagaraju, Y.-C. Toh, S.~Foong, and M.~Hashimoto, ``Crawling, climbing, perching, and flying by fiba soft robots,'' \emph{Science Robotics}, vol.~9, no.~92, p. eadk4533, 2024.
\BIBentrySTDinterwordspacing

\bibitem{overheadperching2}
S.~M. Lee, J.~Liu, J.~L. Chien, W.~H. Ng, M.~Lim, and S.~Foong, ``Rapid resistography with passive overhead-perching mechanism in an unmanned aerial system for wood structure inspection,'' in \emph{2024 IEEE International Conference on Robotics and Automation (ICRA)}, 2024, pp. 1554--1560.

\bibitem{horizon}
\BIBentryALTinterwordspacing
W.~Stewart, L.~Guarino, Y.~Piskarev, and D.~Floreano, ``Passive perching with energy storage for winged aerial robots,'' \emph{Advanced Intelligent Systems}, vol.~5, no.~4, p. 2100150.
\BIBentrySTDinterwordspacing

\bibitem{horizon2}
R.~Zufferey, J.~Tormo-Barbero, D.~Feliu-Talegón, S.~Nekoo, J.~A. Acosta, and A.~Ollero, ``How ornithopters can perch autonomously on a branch,'' \emph{Nature Communications}, vol.~13, 12 2022.

\bibitem{ultra}
A.~McLaren, Z.~Fitzgerald, G.~Gao, and M.~Liarokapis, ``A passive closing, tendon driven, adaptive robot hand for ultra-fast, aerial grasping and perching,'' in \emph{2019 IEEE/RSJ International Conference on Intelligent Robots and Systems (IROS)}, 2019, pp. 5602--5607.

\bibitem{armed}
C.~E. Doyle, J.~J. Bird, T.~A. Isom, C.~J. Johnson, J.~C. Kallman, J.~A. Simpson, R.~J. King, J.~J. Abbott, and M.~A. Minor, ``Avian-inspired passive perching mechanism for robotic rotorcraft,'' in \emph{2011 IEEE/RSJ International Conference on Intelligent Robots and Systems}, 2011, pp. 4975--4980.

\bibitem{tendondriven}
K.~M. Popek, M.~S. Johannes, K.~C. Wolfe, R.~A. Hegeman, J.~M. Hatch, J.~L. Moore, K.~D. Katyal, B.~Y. Yeh, and R.~J. Bamberger, ``Autonomous grasping robotic aerial system for perching (agrasp),'' in \emph{2018 IEEE/RSJ International Conference on Intelligent Robots and Systems (IROS)}, 2018, pp. 1--9.

\bibitem{SNAG}
\BIBentryALTinterwordspacing
W.~R.~T. Roderick, M.~R. Cutkosky, and D.~Lentink, ``Bird-inspired dynamic grasping and perching in arboreal environments,'' \emph{Science Robotics}, vol.~6, no.~61, p. eabj7562, 2021.
\BIBentrySTDinterwordspacing

\bibitem{selfmass}
C.~E. Doyle, J.~J. Bird, T.~A. Isom, J.~C. Kallman, D.~F. Bareiss, D.~J. Dunlop, R.~J. King, J.~J. Abbott, and M.~A. Minor, ``An avian-inspired passive mechanism for quadrotor perching,'' \emph{IEEE/ASME Transactions on Mechatronics}, vol.~18, no.~2, pp. 506--517, 2013.

\bibitem{selfmass2}
C.~E. Doyle, J.~J. Bird, T.~A. Isom, C.~J. Johnson, J.~C. Kallman, J.~A. Simpson, R.~J. King, J.~J. Abbott, and M.~A. Minor, ``Avian-inspired passive perching mechanism for robotic rotorcraft,'' in \emph{2011 IEEE/RSJ International Conference on Intelligent Robots and Systems}, 2011, pp. 4975--4980.

\bibitem{Ozawa}
R.~Ozawa, H.~Kobayashi, and K.~Hashirii, ``Analysis, classification, and design of tendon-driven mechanisms,'' \emph{IEEE Transactions on Robotics}, vol.~30, no.~2, pp. 396--410, 2014.

\bibitem{differential}
\BIBentryALTinterwordspacing
N.~Nikafrooz and A.~Leonessa, ``A single-actuated, cable-driven, and self-contained robotic hand designed for adaptive grasps,'' \emph{Robotics}, vol.~10, no.~4, 2021.
\BIBentrySTDinterwordspacing

\bibitem{j-sugihara}
\BIBentryALTinterwordspacing
J.~Sugihara, M.~Zhao, T.~Nishio, K.~Okada, and M.~Inaba, ``Beatle - self-reconfigurable aerial robot: Design, control and experimental validation,'' 2024. [Online]. Available: \url{https://arxiv.org/abs/2404.09153}
\BIBentrySTDinterwordspacing

\bibitem{eulerbelt}
S.~J. F., \emph{Vector Mechanics for Engineers: Statics}.\hskip 1em plus 0.5em minus 0.4em\relax New York: McGraw-Hill, 1990.

\bibitem{friction}
S.~Blandon, J.~Amaya, and A.~Rojas, ``Development of a 3d printer and a supervision system towards the improvement of physical properties and surface finish of the printed parts,'' pp. 1--7, 10 2015.

\end{thebibliography}


\end{document}